%% file: main.tex
 \documentclass[pmlr,twocolumn,10pt]{jmlr} % W&CP article

\usepackage{booktabs}
\usepackage{siunitx}

\usepackage[switch]{lineno}

\usepackage[utf8]{inputenc} % allow utf-8 input
\usepackage[T1]{fontenc}    % use 8-bit T1 fonts
\usepackage{hyperref}       % hyperlinks
\usepackage{url}            % simple URL typesetting
\usepackage{amsfonts}       % blackboard math symbols
\usepackage{nicefrac}       % compact symbols for 1/2, etc.
\usepackage{microtype}      % microtypography
\usepackage{lipsum}
\usepackage{graphicx}
\usepackage{colortbl}
\usepackage{natbib}
\usepackage{multirow}
\usepackage[nolist,nohyperlinks]{acronym}
\graphicspath{ {./images/} }

\theorembodyfont{\upshape}
\theoremheaderfont{\scshape}
\theorempostheader{:}
\theoremsep{\newline}

\jmlrvolume{259}
\jmlryear{2024}
\jmlrworkshop{Machine Learning for Health (ML4H) 2024} % W&CP title

 \title[MAIRA-Seg: Enhancing Radiology Report Generation with Segmentation-Aware Multimodal LLMs]{MAIRA-Seg: Enhancing Radiology Report Generation with Segmentation-Aware Multimodal Large Language Models}

 \author{%
  \Name{Harshita Sharma} \addr Microsoft Health Futures, Cambridge, UK \Email{harshita.sharma@microsoft.com}\\
  \Name{Valentina Salvatelli \addr Microsoft Health Futures, Cambridge, UK} \Email{vsalvatelli@microsoft.com}\\
  \Name{Shaury Srivastav} \addr Microsoft Research India \Email{t-ssrivastav@microsoft.com}\\
  \Name{Kenza Bouzid} \addr Microsoft Health Futures, Cambridge, UK \Email{kenza.bouzid@microsoft.com}\\
  \Name{Shruthi Bannur} \addr Microsoft Health Futures, Cambridge, UK \Email{shruthi.bannur@microsoft.com}\\
  \Name{Daniel C. Castro} \addr Microsoft Health Futures, Cambridge, UK \Email{dacoelh@microsoft.com}\\
  \Name{Maximilian Ilse} \addr Microsoft Health Futures, Cambridge, UK \Email{maxilse@microsoft.com}\\
  \Name{Sam Bond-Taylor} \addr Microsoft Health Futures, Cambridge, UK \Email{sbondtaylor@microsoft.com}\\
  \Name{Mercy Prasanna Ranjit} \addr Microsoft Research India\Email{mercy.ranjit@microsoft.com}\\
  \Name{Fabian Falck} \addr Microsoft Health Futures, Cambridge, UK \Email{fabian.falck@microsoft.com}\\
  \Name{Fernando Pérez-García} \addr Microsoft Health Futures, Cambridge, UK \Email{fernando.perezgarcia@microsoft.com}\\
  \Name{Anton Schwaighofer} \addr Microsoft Health Futures, Cambridge, UK \Email{antonsc@microsoft.com}\\
  \Name{Hannah Richardson} \addr Microsoft Health Futures, Cambridge, UK \Email{hannah.murfet@microsoft.com}\\
  \Name{Maria Teodora Wetscherek} \addr Microsoft Health Futures, Cambridge, UK; Department of Radiology, University of Cambridge and Cambridge University Hospitals NHS Foundation Trust, Cambridge, UK \Email{a-mwetschere@microsoft.com}\\
    \Name{Stephanie L. Hyland} \addr Microsoft Health Futures, Cambridge, UK \Email{stephanie.hyland@microsoft.com}\\
  \Name{Javier Alvarez-Valle} \addr Microsoft Health Futures, Cambridge, UK \Email{jaalvare@microsoft.com}
 }

\begin{document}
\maketitle
\input{macros}
\input{acronyms}
\thispagestyle{empty}
\begin{abstract}
There is growing interest in applying AI to radiology report generation, particularly for chest X-rays (CXRs).  
This paper investigates whether incorporating pixel-level information through segmentation masks can improve fine-grained image interpretation of multimodal large language models (MLLMs) for radiology report generation. We introduce \textit{MAIRA-Seg}, a segmentation-aware MLLM framework designed to utilize semantic segmentation masks alongside CXRs for generating radiology reports. We train expert segmentation models to obtain mask pseudolabels for radiology-specific structures in CXRs. Subsequently, building on the architectures of MAIRA, a CXR-specialised model for report generation, we integrate a trainable segmentation tokens extractor that leverages these mask pseudolabels, and employ mask-aware prompting to generate draft radiology reports. Our experiments on the \review{publicly available} MIMIC-CXR dataset show that MAIRA-Seg outperforms non-segmentation baselines.  
We also investigate set-of-marks prompting with MAIRA and find that MAIRA-Seg consistently demonstrates comparable or superior performance. 
The results confirm that using segmentation masks enhances the nuanced reasoning of MLLMs, 
potentially contributing to better clinical outcomes.
\end{abstract}


% Version 1
% \paragraph*{Data and Code Availability} We use the publicly available MIMIC-CXR dataset \citep{johnson2019mimic-cxr,johnson2019mimic-cxr-dataset} %hosted on PhysioNet \citep{goldberger2000physiobank}, 
% for the report generation experiments and publicly available CXR segmentation datasets to train segmentation models. More details of the data are described in~\sectionref{sec:data} and ~\sectionref{expert_models}. At present, we are not making the code available.

% Version 2 (shortened)
\paragraph*{Data and Code Availability} We use the  MIMIC-CXR dataset \citep{johnson2019mimic-cxr,johnson2019mimic-cxr-dataset} %hosted on PhysioNet \citep{goldberger2000physiobank}, 
and CXR segmentation datasets which are all publicly available, referring to~\sectionref{expert_models} and ~\sectionref{sec:data} for further details. At present, we do not release the code.

\paragraph*{Institutional Review Board (IRB)}
Proposed use of public datasets was reviewed by home institution. Under policy, use of de-identified public datasets is classified as Not Human Subjects Research [per 45§46.102(e)(1)(ii), 45§46.102(e)(5)]. Guidance and data reflection questions are provided to researchers including considerations to support representativeness, transparency and intended use.

%\vspace{-5pt}
\section{Introduction}
\label{sec:intro}

Radiology report generation involves the automated creation of free-text draft reports from medical images~\citep{liu2019clinicallyaccurate}. Research indicates that applying \ac{AI} in this area could significantly enhance radiology workflows and clinical outcomes~\citep{huang2023generative, yildrim2024multimodal}. There is increasing interest in using \acfp{MLLM} for \acf{CXR} report generation, demonstrating impressive performance~\citep{tu2024medpalmm, chaves2024towards, zhou2024generalist, bannur2024maira2groundedradiologyreport, hyland2024maira1}.
However, current \acp{MLLM} often neglect the integration of pixel-level inputs alongside images for generating  radiology reports, limiting their region-based and fine-grained image interpretation capabilities. %While existing models are capable of developing a global understanding of images, their abilities in region-based or fine-granular image interpretation remain limited. 
This is particularly significant in the biomedical domain, where a single medical image can contain multiple \review{subtle findings}, nuanced structures and relevant context representing the \acp{ROI}. 
% The existing gap provides an opportunity to enhance \ac{MLLM} outputs by introducing fine-grained image information in the form of segmentation masks -- a direction that we aim to investigate in this paper. We hypothesize that providing such localized pixel-level details together with radiology images can enhance the perceptual and reasoning abilities of \acp{MLLM} for biomedical applications like radiology report generation. %, resulting in their improved performance in radiology report generation.
This gap presents an opportunity to enhance \ac{MLLM} outputs by incorporating  segmentation masks, which we aim to explore in this paper. We hypothesize that providing localized pixel-level details alongside images can enhance \ac{MLLM}'s perceptual and reasoning abilities for biomedical applications like radiology report generation.

We propose \textit{MAIRA-Seg}, a segmentation-aware \ac{MLLM} that  utilizes fine-grained mask features from semantic medical image segmentation alongside CXR input images to generate draft radiology reports. We build upon the model architectures and training method from  
the \ac{MAIRA} series of \acp{MLLM}~\citep{bannur2024maira2groundedradiologyreport, hyland2024maira1}. To the best of our knowledge, ours is the first work leveraging semantic image segmentation for instruction tuning \acp{MLLM} for \ac{CXR} report generation. By integrating pixel-level knowledge in the form of segmentation and mask-aware information into the prompt instructions of the \ac{MLLM}, we aim to improve the pixel-wise visual understanding and enhance the quality and accuracy of draft radiology reports generated from \acp{CXR}.

\paragraph{Contributions} Our contributions are: %We make the following contributions.
% \begin{enumerate}
% \setlength{\itemsep}{0pt}
1) We propose MAIRA-Seg, a segmentation aware framework for radiology report generation.
Semantic segmentation masks generated from expert models are integrated into the \ac{MLLM} input using a segmentation tokens extractor, enabling fine-grained supervision along with \ac{CXR} images. %and improving the accuracy of generated draft reports.
2) We train MAIRA-Seg with segmentation masks of multiple anatomical structures, support devices, and pathological regions. By incorporating semantic  segmentation as additional visual inputs along with single or multiple \ac{CXR} views, we achieve superior performance in radiology report generation. We demonstrate notable quantitative and qualitative improvements of MAIRA-Seg over the non-segmentation baseline models.
3) We investigate an additional general-domain method for leveraging segmentation as \ac{MLLM} inputs: \ac{SoM} prompting~\citep{yang2023setofmark}. Here, visual marks (e.g., contours) are directly superimposed on the image for visual instruction tuning. %We perform ablations on the type of \ac{SoM} marks, %and report the best-performing approach, 
For \ac{SoM}, we observe improvements over the non-segmentation baselines and comparable performance to MAIRA-Seg. %However, we observe that the MAIRA-Seg framework outperforms \ac{SoM} prompting on most metrics. %While it achieves 
%\vspace{-5pt}
\section{Related Work}

\paragraph{Multimodal LLMs in Radiology} Recent advances in \ac{AI} for generating free-text radiology reports suggest  improvements in operational efficiency, reduction in radiologist workloads, and enhancement of patient care quality %will be made possible by providing assistance in managing high case volumes, reducing treatment variations, and prioritizing urgent cases needing immediate attention
~\citep{huang2023generative, yildrim2024multimodal, liu2019clinicallyaccurate}.
As a result, there has been a growing research interest in the generation of free-text, narrative-style reports from radiology images~\citep{sloan2024automated}. Specifically, \acp{MLLM} have been increasingly explored and have demonstrated a promising performance for radiology report generation. Recent \acp{MLLM} in the literature encompass generalist biomedical models for multiple imaging modalities~\citep{zhou2024generalist, tu2024medpalmm, yang2024advancing} and specialist radiology domain models \citep{hyland2024maira1, bannur2024maira2groundedradiologyreport, li2024holisticframeworkmultimodallarge, bai2024m3dadvancing3dmedical, chaves2024towards}. For this work, we build on the \ac{CXR}-specific MAIRA framework~\citep{hyland2024maira1, bannur2024maira2groundedradiologyreport}, \review{as it has demonstrated competitive \ac{CXR} report generation performance over prior works such as LLaVA-Rad~\citep{chaves2024towards}, MedPalm-M~\citep{tu2024medpalmm} and MedVersa~\citep{zhou2024generalist}}.
%, namely \maira and \mairatwo~\citep{}. 

%\vspace{-10pt}
\paragraph{Segmentation to Prompt Multimodal LLMs} Recent work in the general domain shows that using \review{visual prompts (e.g. bounding boxes, markers, segmentation masks) with images for \ac{MLLM} visual instruction tuning can enhance their visual perception capabilities%nuanced visual understanding
~\citep{wu2024visualpromptingmultimodallarge}}. Specifically, among methods using segmentation masks, the \acf{SoM} prompting method~\citep{yang2023setofmark} uses off-the-shelf interactive segmentation models to partition the input image into semantically meaningful regions using ``marks'' (e.g., bounding boxes, contours, alphanumeric marks). They query GPT-4V using such marked images and observe that it can provide visually grounded zero-shot responses. Additionally, ~\citet{yan2024listitemsonenew} presents a method to enhance \ac{SoM} prompting ability of existing \acp{MLLM} by using augmented prompts that incorporate mark information. Datasets of \ac{SoM}-augmented images and prompts are curated for fine-tuning MLLMs. We conduct ablations and comparative analysis on \ac{SoM} prompting with MAIRA-Seg and report our key findings. %datasets composed of \ac{SoM}-augmented images and prompts are then curated to fine-tune the \acp{MLLM}. We perform ablations and comparative analysis on~\ac{SoM} prompting with MAIRA-Seg and report our key findings. 
Another approach, the Osprey method~\citep{yuan2024ospreypixelunderstandingvisual}, uses a mask-aware visual extractor to obtain segmentation tokens interleaved with image tokens to prompt the \ac{MLLM}, which is fine-tuned using a curated visual instruction tuning dataset. We adapt the Osprey architecture for the radiology domain (details in \sectionref{sec:methods}). In contrast to Osprey and \ac{SoM}, we generate online mask-aware prompts without the need to generate new instruction tuning datasets for generating the radiology reports. 

For radiology images, recent work~\citep{denner2024visualpromptengineeringmedical} shows that incorporating visual prompts like arrows, circles, and contours for BiomedCLIP models significantly improves lung nodule malignancy classification metrics in \acp{CXR}. Another related study~\citep{zhao2023medicalreportgenerationbased} uses the \ac{SAM}~\citep{kirillov2023segment} to segment meaningful \acp{ROI} of the image and a supervised contrastive learning method showing promising performance for the report generation task. %,  on the IU-XRay public dataset. 
In our knowledge, ours is the first paper leveraging segmentation masks in \acp{MLLM} to improve their nuanced visual understanding for radiology report generation.

%\vspace{-5pt}
\section{Methodology}
\label{sec:methods}

The MAIRA-Seg architecture and method are demonstrated in \figureref{fig:method}. We first train structure-specific expert models for segmenting multiple \ac{CXR} structures. We use these models to generate segmentation masks for the \ac{CXR} images and feed these as pseudolabel inputs to the \ac{MLLM} for training or inference. The masks along with image encoder features are then used to train a \extractor based on the Osprey architecture~\citep{yuan2024ospreypixelunderstandingvisual} that generates two additional segmentation tokens (mask token, spatial token) for each individual mask. We investigate methods to integrate these tokens into the \ac{LLM}'s input, and use interleaved segmentation tokens with text and image tokens. %The segmentation tokens are interleaved with text and image tokens at the input of the \ac{LLM}. 
We augment the input prompts on the fly using the available mask information, without the overhead of curating new instruction tuning datasets to train the \ac{MLLM}.
We describe the individual components of the proposed method in the following sections.
 %and splitting (train, validation, test) are performed  .

\begin{figure*}[t]
\floatconts
  {fig:method}
  {\caption{(a) MAIRA-Seg model architecture. Multi-view and textual context inputs are shown in blue boxes~\citep{bannur2024maira2groundedradiologyreport}. 
  %for frontal and multi-views input (blue boxes). 
  (b) Segmentation tokens extractor architecture based on \citet{yuan2024ospreypixelunderstandingvisual}.}}
  {\includegraphics[width=\linewidth, clip, trim=0cm 0cm 10cm 0cm]{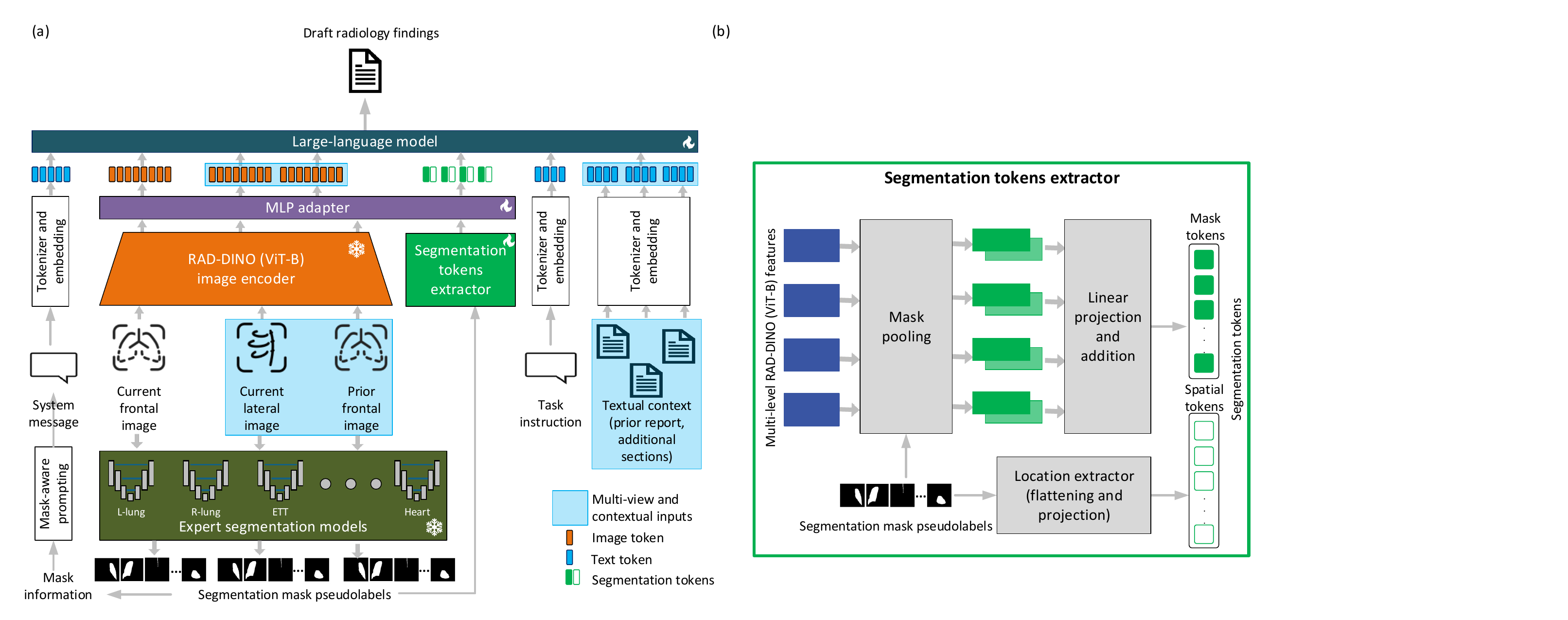}}
\end{figure*}

%\vspace{-5pt}
\subsection{Expert Models for Semantic Medical Image Segmentation}
\label{expert_models}

We leverage expert semantic segmentation models trained for segmenting multiple structures in \acp{CXR}, and predict the corresponding semantic segmentation masks %for multiple structures 
in the input radiology image. \citet{pérezgarcía2024raddinoexploringscalablemedical} empirically evaluate 
multiple segmentation models %such as nnUNet~\citep{isensee2018nnunet}, efficientNet-UNet, BioViL-T with linear head, and \raddino-UPerNet 
for \ac{CXR} structures; the best performance is achieved by EfficientNet-UNet, a U-Net based on the EfficientNet backbone~\citep{tan2020efficientnet}, as such, we use this architecture for our expert models. % We select this U-Net architecture to train our expert models.  

%We train the expert U-Net models for segmenting structures that
For semantic segmentation, we select structures that are relevant to common pathological observations in \acp{CXR} (e.g. CheXpert pathological findings~\citep{chexpert}), and can improve the pixel-level understanding and reasoning capabilities of the trained \ac{MLLM} for \ac{CXR} report generation. We group these structures into three main categories: anatomical, pathological and support devices (\sectionref{app:seg}, \tableref{tab:seg_data}). %We train expert models for each of the nine structures on the full MIMIC-CXR dataset. 
We note that the trained expert semantic segmentation models perform adequately for the respective structures (\sectionref{app:seg}, \tableref{tab:seg_results}).
%report the evaluation performance of the trained expert EfficientNet-UNet semantic segmentation models on the respective structures in \sectionref{app:seg}, \tableref{tab:seg_results}.

There has been an influx of general-purpose, prompt-based (bounding box, points, text) segmentation models in the biomedical domain, for instance, MedSAM~\citep{Ma_2024} and BiomedPARSE~\citep{zhao2024biomedparsebiomedicalfoundationmodel}.
Compared to a U-Net, these prompt-based models require additional input prompts to segment the structures of interest in the \ac{CXR}.
Since we strive for a fully automated system for \ac{CXR} report generation without relying on input prompts, we used the `segment everything' approach of these models, and during our initial experiments, we found that these models were sub-optimal compared to domain- and structure-specific U-Nets.
%results of the `segment everything' modes are noisy and prone to false positives.
%our initial explorations with these models indicated sub-optimal segmentation performance compared to domain- and structure-specific U-Nets; this is also frequently observed in literature~\citep{zhang2023segmentmodelsammedical}. 
%Especially for \acp{CXR}, support devices such as lines and tubes have distinct fine-grained vessel-like appearances where general-purpose models struggle to produce meaningful output segmentation masks. 
Hence, general-purpose, prompt-based segmentation models may not yield precise segmentation for clinically critical tasks like draft radiology report generation.

%\vspace{-5pt}
\subsection{Leveraging Semantic Image Segmentation in \acp{MLLM}}
\label{sec:segmllm}
\paragraph{MAIRA Architecture}
We leverage the MAIRA framework ~\citep{hyland2024maira1, bannur2024maira2groundedradiologyreport} to experiment with semantic segmentation as additional visual inputs along with \ac{CXR} images for generating draft reports. 
MAIRA \acp{MLLM} use a pretrained \ac{CXR}-specific image encoder (\raddino) based on a ViT-B architecture~\citep{pérezgarcía2024raddinoexploringscalablemedical} and a pretrained \ac{LLM} Vicuna-7B v1.5~\citep{vicuna2023}, and align these using a 4-layer \ac{MLP} adapter based on the LLaVA framework~\citep{liu2023llava, liu2023llava15}. The \ac{CXR}-specific \raddino~\citep{pérezgarcía2024raddinoexploringscalablemedical} image encoder is pretrained following the self-supervised DINOv2 approach~\citep{oquab2023dinov2}; it is kept frozen during the \ac{MLLM} training. We use the image encoder checkpoint from~\citet{bannur2024maira2groundedradiologyreport}.
The \ac{LLM} and adapter are fine-tuned in a single stage to generate the draft radiology report. 

\paragraph{MAIRA-Seg Architecture} 
MAIRA-Seg follows the MAIRA architecture for leveraging image and text inputs (\figureref{fig:method}(a)), with some simplifications aimed at reducing the computational needs. 
We replace the \ac{LLM} with Phi-3-mini consisting of 3.8B parameters~\citep{abdin2024phi3technicalreporthighly}, which demonstrated performance close to Vicuna-7B in recent work~\citep{srivastav-etal-2024-maira}. In addition, we do not apply GPT augmentations at training time. Also, we don't perform  grounding in our experiments as in~\citet{bannur2024maira2groundedradiologyreport}. As detailed in \sectionref{sec:data}, the amount of data we train on differs. 

We experiment with two flavours of MAIRA-Seg. A single-view model \textit{\mairasegone} that takes as input only the frontal chest x-ray and their corresponding mask and reports; and a multi-view model \textit{\mairasegtwo} that takes as input also priors and laterals, if available, and the corresponding masks. In the \mairasegtwo model we also incorporate additional `textual context' (prior report, indication, technique and comparison sections of the current report) in the text prompts as  in~\citep{bannur2024maira2groundedradiologyreport}.

%TODO other differences we should highlight here?
In order to leverage the mask inputs, we incorporate into the \ac{MAIRA} framework, a trainable \extractor module based on the Osprey method that uses a visually-aware mask extractor~\citep{yuan2024ospreypixelunderstandingvisual}. The \extractor utilizes the output image features from \raddino (ViT-B architecture)~\cite{pérezgarcía2024raddinoexploringscalablemedical} and individual segmentation masks to generate visually-aware mask and spatial tokens~\citep{yuan2024ospreypixelunderstandingvisual}. A mask token (for one structure) is obtained as a result of mask pooling operation between image and mask features followed by linear projections, addition and \ac{MLP}, to capture mask-level image features. A spatial token is obtained via flattening and linear projection of the segmentation mask only which extracts the  spatial location of the structure of interest. %, capturing its spatial position information. 
We collectively refer to the mask and spatial tokens as `segmentation tokens'. Since this approach involves the extraction of only two additional segmentation tokens per structure, it is scalable to extend to multiple structures and hence masks in the image -- a common scenario found in biomedical images with multiple regions of interest, such as \acp{CXR}. This flexibility makes the approach suitable for fixed computational budgets and context lengths of \acp{LLM}.

A major difference from the Osprey method~\citep{yuan2024ospreypixelunderstandingvisual} is that we use a transformer-based image encoder \raddino~(ViT-B backbone) in contrast to a \ac{CNN}-based CLIP vision encoder used in~\citet{yuan2024ospreypixelunderstandingvisual}. We empirically investigate four design considerations to integrate the the segmentation tokens extractor with MAIRA \acp{MLLM}:
1) the interchangeability of linear and \ac{MLP} layers within the original mask extractor, 2) embedding dimension sizes, 3) layer output indices from the ViT-B based image encoder, and 4) the number of segmentation tokens used to prompt the \ac{LLM}.
%1) interchanging or replacing linear and \ac{MLP} layers in the original mask extractor 2) different embedding dimension sizes 3) different layer output indices to use from the ViT-B based image encoder 4) number of segmentation tokens to prompt the \ac{LLM}. 
We select linear layers for both segmentation tokens, smaller embedding dimension of 768 (opposed to 1024 in the Osprey method), and layer indices [2,4,6,8]  for ViT-B (in contrast to first four layers of the convolutional encoder in Osprey method). We use two segmentation tokens, namely one token for the mask information and one for spatial information (similar to Osprey method), opposed to fewer (1 token) or more (4 or 8 tokens). We propose the architecture of the segmentation tokens extractor as shown in \figureref{fig:method}(b).

%We extend the MAIRA \acp{MLLM} architectures to incorporate the segmentation extractor (\figureref{fig:method}(a)), calling the respective architectures \maira-Seg and \mairatwo-Seg. For the \mairatwo-Seg architecture, similar to \mairatwo architecture, we incorporate prior frontal and current lateral images in addition to the current frontal image, as well as additional `textual context' in the text prompts~\citep{bannur2024maira2groundedradiologyreport}. 
The segmentation tokens extractor, the MLP adapter and \ac{LLM} are all finetuned, while the \raddino image encoder is frozen during the \ac{MLLM} training. More implementation details for MAIRA-Seg models are in~\sectionref{apd:third}.

\paragraph{Tokenization} The segmentation tokens retrieved from the \extractor are used to prompt the \ac{LLM} along with the image and text tokens. We explore multiple approaches to incorporate segmentation tokens as input of the \ac{LLM}, including directly concatenating all image and segmentation tokens, concatenating segmentation tokens in addition to the image tokens, and using separate segmentation tokens for individual structures in the image. We report these in  \tableref{tab:osprey_ablations}. Best outcomes are achieved using separate segmentation tokens for individual structures in the image %and introducing multiple structure-specific segmentation tokens placeholders, 
where the structure-specific tokens interleave with the image and text tokens in the input prompt.  %For the \mairasegone architecture, this leads to a total of nine placeholders for all structures corresponding to the current frontal image (e.g. \texttt{<RLseg>}, \texttt{<LLseg>}, \texttt{<ETTseg>}..), whereas for the \mairasegtwo architecture, it leads to a total of 27 placeholders corresponding to current frontal, prior frontal (e.g. \texttt{<PriorRLseg>}, \texttt{<PriorLLseg>}, \texttt{<PriorETTseg>}..), and current lateral (e.g. \texttt{<LateralRLseg>}, \texttt{<LateralLLseg>}, \texttt{<LateralETTseg>}..) images. 
Every segmentation token is not always used in the text prompt, as the input tokens are only added when a positive segmentation mask is available. Here, positive masks are defined as binary masks with at least one `1' pixel. \review{The rest of the tokenization process follows the method of MAIRA framework for \mairasegone~\citep{hyland2024maira1} and \mairasegtwo~\citep{bannur2024maira2groundedradiologyreport}, respectively.} %An example of the \ac{MLLM} prompt with interleaved visual and text tokens is presented in the following paragraph. 

\paragraph{Mask-aware Prompting} We perform online mask-aware prompting using input mask information, i.e. %where we augment the report generation text prompt with 
the structure names when a positive mask is available, followed by the corresponding segmentation tokens.  This strategy helps us to quickly prototype without the need to curate new instruction tuning datasets to train the \ac{MLLM}. \review{The prompt format and ordering is the following: system message, current frontal image and tokens, positive mask names and tokens, current lateral image tokens,  positive mask names and tokens, prior frontal image, positive mask names and tokens, instruction, textual context (prior report, other
sections). Lateral image, prior image and textual context are added in the \mairasegtwo prompt, similar to ~\citet{bannur2024maira2groundedradiologyreport}}. An example of a text prompt of the \mairasegtwo model for a study with current and prior frontal images and corresponding positive masks is: `You are an expert radiology assistant tasked with interpreting a chest X-ray study. Given the current frontal image\texttt{ <Image>}, left lung mask \texttt{<LLseg>}, right lung mask \texttt{<RLseg>}, endotracheal tube mask \texttt{<ETTseg>}, heart mask \texttt{<Heartseg>} and the prior frontal image \texttt{<PriorImage>}, prior left lung mask \texttt{<priorLLseg>}, prior right lung mask \texttt{<PriorRLseg>}, prior heart mask \texttt{<PriorHeartseg>}, provide a description of the findings in the radiology study in comparison to the prior frontal image. Where segmentation masks are provided to highlight specific image regions...<textual context>'.
%\todo{An example of augmented input prompt for MAIRA-2-Seg is shown in Figure~\ref{}.}. 
% \vspace{-10pt}

%\vspace{-5pt}
\section{Experimental Setup}

We address the task of generating the main body section \findings{} of the text report accompanying a chest X-ray. The task is identical to the radiology report generation task reported in~\citet{hyland2024maira1}. We perform experiments for single-and multi-view MAIRA-Seg and \ac{SoM}, and compare with their respective baselines. 
% We also compare \ac{SoM} prompting with the baselines.
% and its non-grounded version in~\citet{bannur2024maira2groundedradiologyreport}. 

% TODO: We could add here a summary of the experiments in this section once they are all in here

\subsection{Dataset and Evaluation}
\label{sec:data}
We perform the \ac{MLLM} report generation experiments using 
%We train and evaluate MAIRA-Seg and compare with the non-segmentation baselines and the \ac{SoM} prompting method on 
the MIMIC-CXR dataset \citep{johnson2019mimic-cxr,johnson2019mimic-cxr-dataset} hosted on PhysioNet \citep{goldberger2000physiobank}. This dataset from the Beth Israel Deaconess Medical Center in Boston comprises a total of 377,110 DICOM images across 227,835 studies. Each imaging study is accompanied by a report. %We process the DICOM images to remove all non-AP/PA scans. 
For each report, we extract the \findings{} section using the official MIMIC-CXR codebase%
 \footnote{\url{https://github.com/MIT-LCP/mimic-cxr/blob/master/txt/section_parser.py}}. %Data pre-processing is performed following~\cite{hyland2024maira1}. %The official MIMIC-CXR splits~\citep{johnson2019mimic-cxr-dataset} are used.% for our experiments. 
 
 %To train expert segmentation models for generating mask pseudolabels, we utilize publicly available \ac{CXR} datasets, described in more detail in~\sectionref{expert_models}.

We train all models on the training split of MIMIC-CXR and report results on the official MIMIC-CXR test split % i.e. described in~\cite{hyland2024maira1} for \maira-Seg experiments and in~\cite{bannur2024maira2groundedradiologyreport} for \mairatwo-Seg experiments, respectively. 
using standard lexical and clinical metrics~\citep{bannur2024maira2groundedradiologyreport}.   %similar to~\citet{bannur2024maira2groundedradiologyreport}. %on the official MIMIC-CXR test split. 
We also report additional Mask-Relevant (MR) clinical metrics (macro and micro F1-MR) on the CheXpert pathological findings relevant to the segmented structures, namely `Lung Opacity', `Cardiomegaly', `Pneumothorax', `Support Devices', `Pleural Effusion'. These pathological findings are directly correlated with the input segmentation masks (\sectionref{app:seg}, \tableref{tab:seg_data}). It is worth noting that our macro and micro F1-MR are different from the standard macro and micro F1-5~\citep{miura2021improvingfactualcompletenessconsistency} due to the difference in the selected pathological findings for analysis. We select BLEU-4, RadCliQ, \macromr, \micromr and Radfact/logical\_F1 for further analysis in the paper, with additional metrics in Appendix tables. We present the median along with 95\% confidence intervals calculated from 500 bootstrap samples. 
Bold indicates best performance for that metric, or overlapping CIs with best, compared to the baselines. All metrics are higher is better except where `$\downarrow$' indicates lower is better. CheXpert F1 metrics are computed based on CheXbert labeller outputs~\citep{hyland2024maira1}. \review{We also present qualitative results in \figureref{fig:qual,fig:qual1,fig:qual2,fig:qual4}, which were reviewed by a board-certified radiologist.}

\subsection{Baselines and SoM Prompting}
We compare MAIRA-Seg against baseline models, trained solely on input \ac{CXR} images without using semantic segmentation masks, referred to as \textit{\mairaonebaseline} and \textit{\mairatwobaseline}.
%As mentioned earlier, we use a smaller Phi-3 \ac{LLM} to train all \acp{MLLM}.

% Table generated by Excel2LaTeX from sheet 'Sheet7'
\begin{table*}[htbp]
  \centering
  \footnotesize
  \caption{Experimental results for single view and multi-view setup on the official MIMIC-CXR test split. We compare the \ac{SoM} and MAIRA-Seg methods against the non-segmentation baselines. Bold means superior to baselines (i.e., medians do not fall into mutual CIs). F1-MR* are F1 scores on the mask-relevant CheXpert pathological findings, namely, `Lung Opacity', `Cardiomegaly', `Pneumothorax', `Support Devices', `Pleural Effusion'.}
    \centerline{
    \begin{tabular}{@{}p{3.3cm}p{2cm}p{2.2cm}p{2.4cm}p{2.4cm}p{3.1cm}@{}}
    \toprule
     \textbf{Method} & \textbf{BLEU-4} & \textbf{RadCliQ(↓)} & \textbf{ \macromr*} & \textbf{\micromr*} & \textbf{RadFact/logical\_f1} \\
    \midrule
%  Single view   
      \textbf{\mairaonebaseline} & 
      14.2 [13.7, 14.8]  & 3.19 [3.15, 3.22] & 54.1 [51.2, 56.2] & 61.4 [60.0, 62.7]  & 42.4 [41.5, 43.6] \\
      \textbf{\somone} & 
      14.6 [14.1, 15.2] & \textbf{3.14} [3.10, 3.18] & 55.3 [53.1, 57.8] & \textbf{63.1} [61.9, 64.4] & \textbf{44.0} [43.0, 45.1]  \\
      \textbf{\mairasegone} & 
      14.5 [14.0, 15.1] & \textbf{3.11} [3.08, 3.15] & \textbf{59.2} [56.7, 61.7] & \textbf{65.4} [64.1, 66.7] & \textbf{44.7} [43.8, 45.8] \\
      \cmidrule{1-6}
%  Multiple views
      \textbf{\mairatwobaseline} & 
      19.5 [19.0, 20.1] & 2.90 [2.86, 2.94]  & 52.5 [50.0, 55.0]  & 55.6 [54.2, 57.0] & 45.4 [44.2, 46.5]  \\
      \textbf{\somtwo} & 
      20.3 [19.8, 20.8] & \textbf{2.81} [2.77, 2.85]  &  53.9 [51.5, 56.3] &  \textbf{59.9} [58.6, 61.1] &  \textbf{47.4} [46.3, 48.6] \\
      \textbf{\mairasegtwo} & 
      19.6 [19.1, 20.1] & \textbf{2.82} [2.78, 2.86] & 55.9 [53.6, 58.7] & \textbf{60.9} [59.6, 62.5] & \textbf{47.1} [46.1, 48.3] \\
    \bottomrule
    \end{tabular}%
    }%
  \label{tab:maira_test}%
\end{table*}%

\begin{figure*}[!h]
 % Caption and label go in the first argument and the figure contents
 % go in the second argument
\floatconts
  {fig:stratified}
  {\caption{Stratified F1-scores for the five mask-relevant findings comparing the respective baselines, \ac{SoM} prompting and MAIRA-Seg models for (a) single view (b) multi-view experiments (c) Support for each pathological finding in the MIMIC-CXR test set.}} %\todo{Update stratified results for MAIRA-2-Seg}.}}
  {\includegraphics[width=\linewidth]{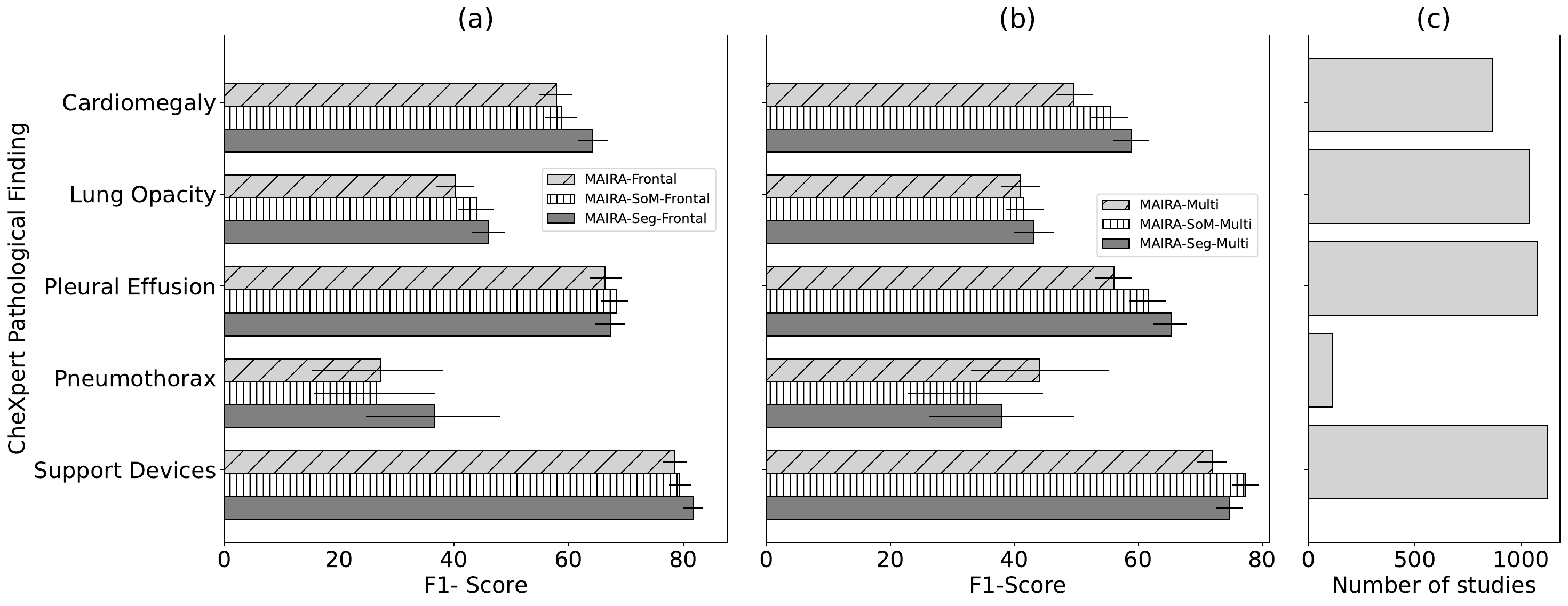}}
\end{figure*}

Moreover, we explore the \acf{SoM} prompting method~\citep{yang2023setofmark, yang2024advancing} that uses ``marks'' derived from segmentation overlaid on input images to visually prompt the \acp{MLLM} and has shown promising results in the general domain. % \ac{SOTA} approach , namely, . 
For this purpose, we create grayscale contours and alphanumeric marks using the segmentation mask pseudolabels and overlay these on the \ac{CXR} images \review{(example in \sectionref{ablation_som}, \figureref{fig:somexample})}. 
We demonstrate the usability of the \ac{SoM} prompting for the radiology report generation task through ablations on contours and alphanumeric marks, and the plain prompt and augmented prompt approaches (\sectionref{ablation_som}, \tableref{tab:somablations}). For the augmented prompts, we use the online mask-aware augmentations as explained in~\sectionref{sec:segmllm}. We also list all masks sequentially with the corresponding mark numbers at the end of the prompt~\citep{yang2024advancing}. We find that using contours and alphanumeric marks with augmented prompts performed the best in our use-case (\sectionref{ablation_som}, \tableref{tab:somablations}). %We use this configuration for comparison experiments with MAIRA-Seg \acp{MLLM}. For the MAIRA-Seg architecture we experiment with both single view (\textit{\somone}) and multi-view inputs (\textit{\somtwo}).

%\vspace{-10pt}
\section{Results}
We report the results for the single view and multi-view inputs in \tableref{tab:maira_test}, and extended tables with additional lexical and clinical metrics in \sectionref{sec:detailed-results}, \tableref{tab:maira1_test,tab:maira2_test} to supplement the key results. %The results for the multi-view experiments are depicted in \tableref{tab:maira2_test}. 
We also present the stratified F1-score for the CheXpert MR pathological findings, along with support for each multi-label finding in \figureref{fig:stratified} (extended 14 CheXpert pathologies in \sectionref{sec:detailed-results}, \figureref{fig:stratified_all}). %We highlight the findings relevant to the segmented structures. 
Results for ablations are presented in \tableref{tab:osprey_ablations}. 
We present one qualitative example %of the input \ac{CXR} image(s), target and predicted report 
in \figureref{fig:qual}, with more examples in \sectionref{app:qual}. %, with some examples in the following sections.

\subsection{MAIRA-Seg-Frontal}
\label{sec:results_frontal}
\paragraph{Quantitative Analysis}
%In \sectionref{sec:detailed-results}, \tableref{tab:maira1_test}, we provide additional lexical and clinical metrics to supplement \tableref{tab:maira_test}. 
From \tableref{tab:maira_test}, We observe that for the single view experiments, leveraging segmentation along with input images to prompt the \ac{MLLM} improves report generation performance. \mairaonebaseline is outperformed by \mairasegone in all the clinical metrics. We observe comparable results for \somone, with significant improvements over baseline in 3 out of 4 clinical metrics, but \mairasegone is superior for extended clinical metrics as shown in~\tableref{tab:maira1_test}. For the BLEU-4 lexical metric, we do not find significant differences, which aligns with our expectation since we aim at enhancing visual understanding capabilities of the \ac{MLLM}, indicated by the clinical metrics. 
%For the remaining metrics, medians are always higher, however, \acp{CI} overlap with the baseline.
%We observe significant gains for clinical metrics (RG\_ER, CheXbert vector, RadCliQ, and 3 out of 4 F1-scores) for \maira-Seg. The overall metric medians are higher by large margins, even when some metrics have more overlapping \acp{CI} with the baseline (RadGraph-F1 anf MacroF1-14). For the RadFact metrics, we find significant improvements from baseline on recall and F1-score. 
When further investigating the F1-MR stratified scores, %  by CheXpert pathological finding
in~\figureref{fig:stratified}(a), we find that  %namely, support devices, pneumothorax, pleural effusion, lung opacity and cardiomegaly, 
\mairasegone outperforms \mairaonebaseline on all five mask-relevant pathological findings, with significant gains in support devices, lung opacity and cardiomegaly.   %\somone also shows an improvement for these findings, and is superior to \mairasegone for pleural effusion.

\paragraph{Qualitative Analysis} \figureref{fig:qual} reveals that \mairaonebaseline omits the detail of lungs being ``hyperinflated'' in the draft report, however, this is correctly predicted by \mairasegone, probably due to enhanced visual understanding using the existing lung masks. Interestingly, in \sectionref{apd:second}, \figureref{fig:qual1}(b)-(c), we find that the tip locations of tubes are correctly predicted by \mairasegone in contrast to \mairaonebaseline, which shows the importance of adding segmentation masks for fine-grained tubular structures. Moreover, we find that the accuracy of masks affects the generated report outcomes, reflected in \sectionref{apd:second}, \figureref{fig:qual2}(a) where overlapping devices lead to over-segmentation and predicting ``distal SVC'' in place of ``mid SVC'' as the tip location, suggesting that more accurate segmentation masks can lead to more precise report generation.

\begin{figure*}[!h]
 % Caption and label go in the first argument and the figure contents
 % go in the second argument
\floatconts
  {fig:qual}
  {\vspace{-2em}\caption{Qualitative result for an example in the MIMIC-CXR test set, showing target and predicted \textit{Findings} using \mairaonebaseline and  \mairasegone. Mask pseudolabels are shown overlaid on the \ac{CXR} image for illustrative purposes (corresponding masks are used to obtain segmentation tokens).}}
  {\includegraphics[width=\linewidth, clip, trim=20pt 9cm 10pt 20pt]{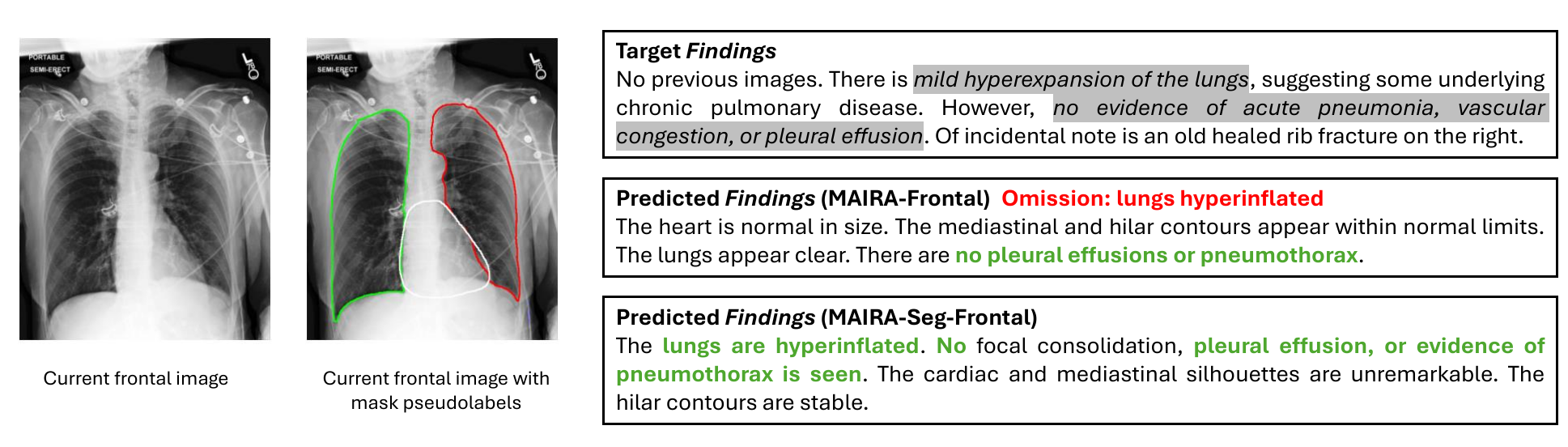}}
\end{figure*}

%Hence, the improvements that were observed over non-segmentation baseline on the MIMIC-CXR validation set for \maira experiments (\sectionref{ablations_concat}, \tableref{tab:osprey_ablations}) can also be seen for the test split, validating our hypothesis that leveraging segmentation can improve the \ac{MLLM} performance for report generation.

\paragraph{Ablations}
\label{ablations_concat}
\begin{table*}[!h]
\footnotesize
  \centering
  \caption{Ablation of strategies for incorporating segmentation tokens on MIMIC-CXR validation split. (DC: direct concatenation, CS: combined segmentation tokens, SS: structure-specific segmentation tokens, NS: no segmentation tokens).}
    \centerline{
    \begin{tabular}{lrrrrrr}
    \toprule
    \textbf{Method} & \multicolumn{1}{p{75pt}}{\textbf{\mairaonebaseline}} & \multicolumn{1}{p{50pt}}{\textbf{Ablation 1}} & \multicolumn{1}{p{50pt}}{\textbf{Ablation 2}} & \multicolumn{1}{p{50pt}}{\textbf{Ablation 3}} & \multicolumn{1}{p{50pt}}{\textbf{Ablation 4}} & \multicolumn{1}{p{50pt}}{\textbf{Ablation 5}} \\
    \midrule
    \multicolumn{1}{p{7.285em}}{\textbf{Concatenation strategy}} & N/A   & \multicolumn{1}{r}{ SS} & \multicolumn{1}{r}{NS} & \multicolumn{1}{r}{DC} & \multicolumn{1}{r}{ CS} & \multicolumn{1}{r}{SS} \\
    \midrule
    \multicolumn{1}{p{7.285em}}
    {\textbf{Augmented mask-aware prompts?}} & No    & No    & Yes   & Yes   & Yes   & Yes \\
    \midrule
    %\multicolumn{6}{l}{\textit{\textbf{Lexical }}} &  \\
    BLEU-4  & 17.5  & 18.2  & 18.4  & 17.9  & \multicolumn{1}{r}{18.4} & \textbf{18.7} \\
    %\midrule
    %\multicolumn{6}{l}{\textit{\textbf{Clinical }}} &  \\
    RadCliQ (↓) & 2.72 [2.70, 2.75] &  2.75 [2.73, 2.78] & 2.72 [2.70, 2.75] & 2.75 [2.72, 2.78] & 2.73 [2.70, 2.75] & \textbf{2.71 }[2.68, 2.73] \\
    \macrofourteen  & 33.8 [32.8, 34.8] & 34.1 [33.1, 35.3] & 33.6 [32.7, 34.5] & 33.4 [32.3, 34.4] & 35.2 [34.3, 36.2] & \textbf{37.0} [35.8, 38.0] \\
    \microfourteen  & 51.9 [51.0, 52.7] & 51.6 [50.8, 52.4] & 52.2 [51.4, 53.0] & 50.8 [50.0, 51.6] & 53.1 [52.3, 53.9] & \textbf{54.0} [53.2, 54.8] \\
    \bottomrule
    \end{tabular}%
    }
  \label{tab:osprey_ablations}%
 % \vspace{-1em}
\end{table*}%

 We explore multiple approaches to incorporate segmentation tokens as input of the \ac{LLM}. These include 1) concatenating image and segmentation features in unified image tokens, where the image and text tokens are interleaved (DC: direct concatenation); 2) concatenating segmentation tokens from all the structures in the image, where the combined segmentation tokens are interleaved with the image and text tokens (CS: Combined segmentation tokens); 3) using separate segmentation tokens for individual structures in the image, where the structure-specific tokens interleave with the image and text tokens  (SS: structure-specific tokens). Results are depicted in \tableref{tab:osprey_ablations} for the MIMIC-CXR validation split. We observe that the third strategy (SS) gives the best results  and we select this strategy for introducing segmentation tokens at the input of the \ac{LLM}. We also perform ablations on the impact of segmentation tokens and mask-aware prompting (\tableref{tab:osprey_ablations}, third and fourth columns), and observe improvements from the baseline for both cases, highlighting the importance of both sources of information to train the \ac{MLLM}.

\subsection{MAIRA-Seg-Multi}

\paragraph{Quantitative Analysis}
%In \sectionref{sec:detailed-results}, \tableref{tab:maira2_test}, we provide additional lexical and clinical metrics to supplement \tableref{tab:maira_test} for the multi-view experiments. 
In \tableref{tab:maira_test}, we observe significant improvements over the \mairatwobaseline for 3 out of 4 clinical metrics using both \somtwo or \mairasegtwo to leverage segmentation in \acp{MLLM}. Macro F1-MR improvements are not significant in both cases, which could be attributed to data imbalance.

We find \somtwo is closer in performance to \mairasegtwo, in contrast to the single-view experiments. This can be due to the fact that our architecture choices are based on ablation and tuning experiments using the single view counterparts (fewer input tokens to the \ac{LLM}). Also, using a larger number of input tokens in \mairasegtwo with the same-sized LLM could lead to more complex interactions, unlike \somtwo, which maintains a constant number of image tokens. Thus, we report \mairasegtwo results as proof-of-concept with significant gains over \mairatwobaseline and consistent findings with \mairasegone, however, there is potential for enhancing the \mairasegtwo architecture for optimal performance which we leave for future work.
%. Again, all clinical and lexical metrics show improvements  using either \ac{SoM} or \mairatwo-Seg that leverage segmentation in \acp{MLLM}. For \mairatwo-Seg, we observe significant gains %for 3 out of 4 lexical metrics and 
% for RG\_ER, CheXbert vector, RadCliQ, Micro-F1-14 and Micro-F1-MR clinical metrics. Moreover, we see improvements in the overall median for clinical metrics such as MacroF1-14, Macro and macro F1-MR by large differences however the \acp{CI} are overlapping with the baseline. RadFact metrics show improved medians from the baseline, with significant gains in logical precision for both \ac{SoM} prompting and \maira-Seg. 
For the pathology-stratified F1-MR scores in~\figureref{fig:stratified}(b), we observe significant improvements for \mairasegtwo from the non-segmentation baseline for most relevant pathological findings such as support devices, cardiomegaly and pleural effusion.
%, scores are better for lung opacity, and there is an insignificant drop for pneumothorax.
%For \mairatwo-Seg, we observe drops for enlarged cardiomediastinum and lung lesion from the non-segmentation baseline, however, the support for these findings is very low in the test set to make concrete conclusions.
%More examples and detailed description for MAIRA-1-Seg are presented in \figureref{fig:qual1}(b), and \figureref{fig:qual3}(a)-(b). 

\paragraph{Qualitative Analysis}
Observing the qualitative results, in \sectionref{apd:second}, \figureref{fig:qual4}(a), \mairasegtwo and \mairatwobaseline correctly mention stable cardiomegaly, bilateral pleural effusions and left retrocardiac opacity. However, \mairatwobaseline hallucinates a right-sided PICC line, which is not mentioned by \mairasegtwo, also not detected as a mask by the expert segmentation model %as a \ac{CVC} mask 
in current or prior images. Moreover, \mairatwobaseline hallucinates  edema. %On the other hand, 
Improvement of aeration is omitted in both predicted draft reports.  

\subsection{Set-of-Marks Prompting}
Although \ac{SoM} prompting shows competitive results and also outperform the non-segmentation baselines in several instances, we noted several challenges with such a prompting method for biomedical images. Firstly, as multiple masks are overlaid on the image in the form of alphanumeric characters and/or contours, these may lead to undesirable occlusions in the image -- while this may not be problematic in natural scenes, it could be consequential for biomedical images containing small structures (e.g. rib fractures or lung lesions in \ac{CXR} images). Moreover, as we scale-up to multiple structures in a single image, it may increase the number of occlusions and lead to confusions in the attended image features by the~\ac{MLLM} (e.g. multiple overlapping lines, overlapping anatomical regions with tubes). In contrast, it is easier to scale up MAIRA-Seg to more structures, which is only limited by the context length of the \ac{LLM}. \review{Further, as \ac{CXR} images are grayscale, there are limited options available for contour intensities to appear with reasonable contrast (\sectionref{ablation_som}, \figureref{fig:somexample}) compared to RGB contours frequently used in the general domain. } Lastly, it is unclear if the frozen \raddino image encoder~\citep{pérezgarcía2024raddinoexploringscalablemedical} trained on \acp{CXR} can interpret out-of-domain images with the set of marks. %, since it was not trained with such marked \ac{CXR} images.

%\todo{Add more qualitative results for MAIRA-2?}
\vspace{-10pt}
\section{Conclusions and Future Work}

We presented \textit{MAIRA-Seg}, a proof-of-concept \ac{MLLM} framework to leverage semantic segmentation masks for \ac{CXR} radiology report generation. Our experiments demonstrate improvements in clinical metrics (with additional mask-relevant metrics) using the proposed approach on MAIRA architectures. We observe encouraging improvements in quantitative results for MAIRA-Seg when introducing segmentation masks with \ac{CXR} images to visually prompt the \ac{MLLM}. This confirms that segmentation can enhance the nuanced fine-granular reasoning and understanding of MAIRA-Seg to interpret complex biomedical images. We compare our method  %the set-of-marks prompting approach 
for single- and multi-view baseline models, where we outperform the latter in most cases. However, set-of-marks also achieves competitive metrics over the non-segmentation baselines, reconfirming our hypothesis that semantic segmentation can potentially lead to better \ac{MLLM} understanding of radiological images and generation of superior draft reports. 

\review{The proposed MAIRA-Seg MLLM framework is demonstrated on \ac{CXR} report generation, however, due to the flexibility in its architecture, it can be  extended to other imaging modalities such as CT and MRI, given a modality-specific image encoder, expert segmentation models and modality-specific training report generation data.} In future, we intend to explore ways to improve the segmentation quality of the generated mask pseudo-labels (e.g. there may be under or over-segmentation), especially for the fine-grained structures. Moreover, we will explore more generalist segmentation models in our framework in order to scale up to multiple unseen structures without sacrificing segmentation performance. \review{Through model optimization and tuning methods, specifically for the \mairasegtwo model, we will aim to further improve model performance (for instance, by using more training datasets and extending to grounded reporting) and optimize computational requirements at inference time (for instance, quantization and pruning), with the end goal of clinical deployment. We intend to perform more elaborate assessment of results through human-centric evaluation, similar to the study described  in~\citet{bannur2024maira2groundedradiologyreport}.} Lastly, we aim to continue exploring the interactions between the visual (images, segmentation) and text tokens in \acp{MLLM}. %Additionally, we want to explore ways of measuring uncertainty of outputs from the \ac{LLM} for this problem, aligned with the ongoing field of research.

\vfill
\bibliography{references}

\appendix

\section{More Experimental Details}\label{apd:first}
To utilize the available computational requirements economically for fixed computational budgets, we perform ablations and tuning experiments using the \mairasegone architecture  which uses only frontal images at the input. We then transfer key findings to \mairasegtwo with multi-view inputs including current lateral image, prior image, additional report sections (see \figureref{fig:method}(a)). We report our ablation results on the MIMIC-CXR validation split~\citep{bannur2024maira2groundedradiologyreport}. 

\subsection{Ablations: Set-of-marks Prompting Method}
\label{ablation_som}

\begin{figure}
    \centering
    \includegraphics[width=0.6\linewidth]{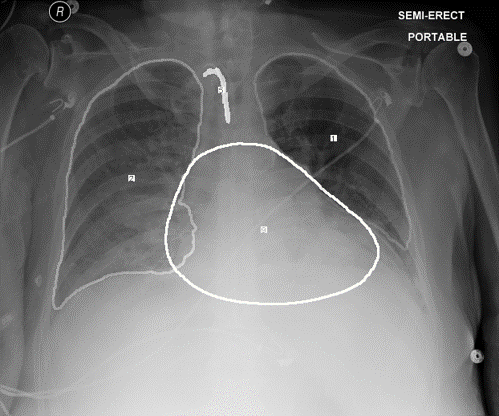}
    \caption{Example of a \ac{CXR} image overlaid with set-of-marks for the SoM prompting method.}
    \label{fig:somexample}
\end{figure}

\review{Example of a \ac{CXR} image overlaid with set-of-marks is presented in \figureref{fig:somexample}}. Results of ablations on set-of-marks prompting are depicted in \tableref{tab:somablations} for the MIMIC-CXR validation split. We observe that mask-aware augmented prompts help compared to settings without prompt augmentation. Also, presence of contours and alphanumeric marks are more beneficial than just contours, consistent with the observations in~\citep{yang2023setofmark}. We also performed preliminary sanity checks where we used the same grayscale intensity for contours, or the same alphanumeric mark, for different masks, however these settings lead to worse performance compared to the classic \ac{SoM} method.

% Table generated by Excel2LaTeX from sheet 'Sheet4'
\begin{table*}[htbp]
\footnotesize
  \centering
  \caption{Ablations of set-of-marks prompting method on MIMIC-CXR validation split.}
    \begin{tabular}{p{65pt}rrrrr}
    \toprule
    \textbf{Method} & \textbf{\mairaonebaseline} & \textbf{Ablation 1} & \textbf{Ablation 2} & \textbf{Ablation 3} & \textbf{Ablation 4} \\
    \midrule
    \textbf{Contours?} & No    & Yes   & Yes   & Yes & Yes \\
    \textbf{Alphanumeric marks?} & No    & No    & No & Yes & Yes \\
    \textbf{Augmented mask-aware prompts?} & No    & No    & Yes   & No    & Yes \\
    \midrule
    \multicolumn{6}{l}{\textit{\textbf{Lexical }}} \\
    ROUGE-L & 34.8 [34.4, 35.3] & 34.5 [34.0, 34.9] & 35.0 [34.6, 35.4] & 34.9 [34.4, 35.3] & 35.3 [34.9, 35.7] \\
    BLEU-1 & \multicolumn{1}{r}{38.5} & \multicolumn{1}{r}{37.6} & \multicolumn{1}{r}{38.2} & \multicolumn{1}{r}{37.9} & \multicolumn{1}{r}{\textbf{38.9}} \\
    BLEU-4  & \multicolumn{1}{r}{17.5} & \multicolumn{1}{r}{17.1} & \multicolumn{1}{r}{17.5} & \multicolumn{1}{r}{17.3} & \multicolumn{1}{r}{\textbf{17.8}} \\
    METEOR & \multicolumn{1}{r}{37.5} & \multicolumn{1}{r}{36.8} & \multicolumn{1}{r}{37.4} & \multicolumn{1}{r}{37.2} & \multicolumn{1}{r}{\textbf{37.9}} \\
    \midrule
    \multicolumn{6}{l}{\textit{\textbf{Clinical }}} \\
    RadGraph-F1 & 30.6 [30.1, 31.2] & 30.0 [29.4, 30.5] & 30.8 [30.3, 31.3] & 30.4 [30.0, 30.9] & 31.5 [30.9, 32.0] \\
    \rger  & 35.0 [34.5, 35.5] & 34.3 [33.8, 34.8] & 35.1 [34.6, 35.6] & 34.8 [34.3, 35.3] & \textbf{35.8} [35.4, 36.3] \\
    CheXbert vector & 51.5 [50.9, 52.1] & 50.7 [50.1, 51.3] & 51.7 [51.1, 52.3] & 51.0 [50.5, 51.6] & 52.4 [51.7, 52.9] \\
    RadCliQ (↓) & 2.72 [2.70, 2.75] & 2.76 [2.73, 2.79] & 2.72 [2.69, 2.74] & 2.74 [2.71, 2.76] & \textbf{2.68 }[2.66, 2.71] \\
    \macrofourteen  & 33.8 [32.8, 34.8] & 32.5 [31.6, 33.5] & 34.9 [33.9, 36.1] & 33.6 [32.7, 34.6] & \textbf{36.3} [35.2, 37.4] \\
    \microfourteen  & 51.9 [51.0, 52.7] & 49.7 [48.8, 50.4] & 52.2 [51.4, 53.0] & 51.5 [50.7, 52.4] & \textbf{53.6} [52.8, 54.4] \\
    \bottomrule
    \end{tabular}%
  \label{tab:somablations}%
\end{table*}%

\subsection{Expert Semantic Segmentation Models}
\label{app:seg}

% Table generated by Excel2LaTeX from sheet 'Sheet4'
\begin{table*}[htbp]
  \centering
  \footnotesize
  \caption{Dataset details for training expert segmentation models and associated CheXpert findings classes.}
    \centerline{
    \begin{tabular}{@{}p{5em}lp{15.07em}rp{13.855em}@{}}
    \toprule
    \textbf{Category} & \multicolumn{1}{p{6.5em}}{\textbf{Structure}} & \textbf{Training dataset} & \multicolumn{1}{p{5em}}{\textbf{Number of images}} & \textbf{Correlated CheXpert findings~\citep{chexpert}} \\
    \midrule
    \multirow{3}[2]{*}{Anatomical} & Left lung &  \multirow{2}[1]{*}{MIMIC-CXR subset~\citep{chen2022cxrmimic}} & \multirow{2}[1]{*}{1,138} & \multirow{2}[1]{*}{Lung opacity} \\
    \multicolumn{1}{c}{} & Right lung & \multicolumn{1}{l}{} &       & \multicolumn{1}{l}{} \\
    \cmidrule{2-5}
    \multicolumn{1}{c}{} & Heart & CheXmask~\citep{Gaggion_2024} & ~242,000 & Cardiomegaly \\
    \midrule
    \multirow{6}[2]{5em}{Support devices} & \ac{CVC}  & \multirow{4}[1]{*}{RANZCR-CLiP~\citep{tang2021clip}} & \multirow{4}[1]{*}{30,083} & \multicolumn{1}{l}{\multirow{4}[1]{*}{Support devices}} \\
    \multicolumn{1}{l}{} & \ac{ETT}   & \multicolumn{1}{l}{} &       & \multicolumn{1}{l}{} \\
    \multicolumn{1}{l}{} & \ac{NGT}  & \multicolumn{1}{l}{} &       & \multicolumn{1}{l}{} \\
    \multicolumn{1}{l}{} & \ac{SGC}   & \multicolumn{1}{l}{} &       & \multicolumn{1}{l}{} \\
    \cmidrule{2-5}
    \multicolumn{1}{l}{} & Chest tube & CANDID-PTX~\citep{feng2021curation} & 19,237 & Support devices, Pneumothorax, Pleural effusion \\
    \midrule
    Pathological & Pneumothorax & CANDID-PTX~\citep{feng2021curation} & 19,237 & Pneumothorax \\
    \bottomrule
    \end{tabular}%
    }
  \label{tab:seg_data}%
\end{table*}%

 %Some of these results have been reported previously in~\citep{pérezgarcía2024raddinoexploringscalablemedical}
We present the datasets used for training expert EfficientNet-UNet semantic segmentation models in \tableref{tab:seg_data} and report segmentation metrics in \tableref{tab:seg_results}. For the latter, we aggregate the scores only over the positives and provide mean and standard deviation across the test set (70/15/15 split over each dataset in~\tableref{tab:seg_data}). CL-Dice metric~\citep{Shit_2021} is additionally reported to measure segmentation performance on tubular structures in the support devices category. \review{Implementation details including segmentation preprocessing are provided in \sectionref{apd:third}.}

% Table generated by Excel2LaTeX from sheet 'Sheet4'
\begin{table}[htbp]
\footnotesize
  \centering
  \caption{Segmentation metrics for the expert EfficientNetUNet semantic segmentation models. `Supp.' stands for support. }%~\citep{pérezgarcía2024raddinoexploringscalablemedical}}
    \begin{tabular}{p{4.5em}p{5em}rl}
    \toprule
    \textbf{Category} & \textbf{Structure} & \multicolumn{1}{l}{\textbf{Dice score}} & \textbf{CL-Dice} \\
    \midrule
    \multirow{3}{*}{Anatomical} & Left lung &    \multicolumn{1}{l}{98.3 $\pm$ 1.1 }    & N/A \\
          & Right lung &  \multicolumn{1}{l}{98.3 $\pm$ 1.4 }      & N/A \\
          & Heart & \multicolumn{1}{l}{95.3 $\pm$ 2.0 }      & N/A \\
    \midrule
    \multirow{5}{*}{Supp.devices} & CVC   & \multicolumn{1}{l}{77.0 $\pm$ 13.0 } & 87.5 $\pm$ 15.0  \\
          & ETT   & \multicolumn{1}{l}{58.5 $\pm$ 24.1 } & 66.3 ± 32.8  \\
          & NGT   & \multicolumn{1}{l}{69.5 $\pm$17.0} & 79.6 ± 20.9  \\
          & SGC   & \multicolumn{1}{l}{73.2 $\pm$ 12.6} & 84.9 ± 15.5  \\
          & Chest tube & \multicolumn{1}{l}{53.0 $\pm$ 16.8} & 59.5 $\pm$ 20.8 \\
    \midrule
    Pathological & Pneumothorax & \multicolumn{1}{l}{73.5 $\pm$ 26.9}     & N/A \\
    \bottomrule
    \end{tabular}%
  \label{tab:seg_results}%
\end{table}%

\subsection{Detailed Results}\label{sec:detailed-results}

We present the experimental results for \mairasegone and \mairasegtwo with additional standard lexical and clinical metrics for radiology report generation in \tableref{tab:maira1_test} and \tableref{tab:maira2_test}. We illustrate stratified F1-scores for 14 CheXpert pathological findings in \figureref{fig:stratified_all}.

\begin{table*}[!h]
  \centering
  \footnotesize
  \caption{Extended single view experimental results on the official MIMIC-CXR test split and comparison of the segmentation-aware methods \somone and \mairasegone against the \mairaonebaseline baseline.} %(F1-MR* are F1 scores on the mask-relevant CheXpert pathological findings, namely, `Lung Opacity', `Cardiomegaly', `Pneumothorax', `Support Devices', `Pleural Effusion').}
    \begin{tabular}{lp{3cm}rr}
    \toprule
    \textbf{Method} & \multicolumn{1}{l}{\textbf{\mairaonebaseline}} & \multicolumn{1}{l}{\textbf{\somone}} & \multicolumn{1}{l}{\textbf{\mairasegone}} \\
    \midrule
    \textit{\textbf{Lexical }} &        &       &  \\
    ROUGE-L &  29.3 [28.8, 29.9] & 30.0 [29.5, 30.5] & 29.8 [29.3, 30.3] \\
    BLEU-1  & 36.0 [35.3, 36.9] & 36.7 [35.9, 37.4] & 37.1 [36.4, 37.8] \\
    BLEU-4 & 14.2 [13.7, 14.8]  & 14.6 [14.1, 15.2] & 14.5 [14.0, 15.1] \\
    METEOR & 31.2 [30.7, 31.9] & 31.9 [31.3, 32.4]  & 32.0 [31.5, 32.5] \\
    \midrule
    \textit{\textbf{Clinical }} &       &       &  \\
    RadGraph-F1 & 23.4 [22.9, 24.1] & 24.4 [23.8, 25.0] & 24.5 [23.8, 25.0] \\
    \rger & 28.1 [27.4, 28.8] & \textbf{29.0} [28.4, 29.6] & \textbf{29.3} [28.7, 29.9] \\
    CheXbert vector &  41.7 [40.8, 42.5] & 42.3 [41.5, 43.2] & \textbf{43.4} [42.5, 44.3] \\
    RadCliQ (↓) &  3.19 [3.15, 3.22] & \textbf{3.14} [3.10, 3.18] & \textbf{3.11} [3.08, 3.15] \\
    \macrofourteen  & 31.9 [30.3, 33.4] & 32.6 [31.0, 33.9] & \textbf{34.5} [32.9, 36.0] \\
    \microfourteen  & 50.4 [49.2, 51.7] & 51.6 [50.5, 52.9] & \textbf{53.5 }[52.3, 54.6] \\
   \macromr* & 54.1 [51.2, 56.2] & 55.3 [53.1, 57.8] & \textbf{59.2} [56.7, 61.7] \\
   \micromr* & 61.4 [60.0, 62.7] & \textbf{63.1} [61.9, 64.4] & \textbf{65.4} [64.1, 66.7] \\
    RadFact/logical\_precision &  46.8 [45.5, 48.0] & \textbf{48.3} [47.0, 49.3] & \textbf{48.4} [47.3, 49.6] \\
    RadFact/logical\_recall & 38.9 [37.8, 40.1] & \textbf{40.5} [39.4, 41.6] & \textbf{41.5} [40.5, 42.7] \\
    RadFact/logical\_f1 & 42.4 [41.5, 43.6] & \textbf{44.0} [43.0, 45.1] & \textbf{44.7} [43.8, 45.8] \\
    \bottomrule
    \end{tabular}%
  \label{tab:maira1_test}%
\end{table*}%

\begin{table*}[!h]
  \centering
  \footnotesize
  \caption{Extended multi-view experimental results for \mairasegtwo on the official MIMIC-CXR test split and comparison of the segmentation-aware methods \somtwo and \mairasegtwo against the \mairatwobaseline baseline.}
    \begin{tabular}{lp{3cm}p{3cm}p{3cm}}
    \toprule
    \multicolumn{1}{l}{\textbf{Method}} &\textbf{\mairatwobaseline} & \textbf{\somtwo} & \textbf{\mairasegtwo} \\
    \midrule
    \multicolumn{4}{l}{\textit{\textbf{Lexical }}} \\
    ROUGE-L & 35.5 [34.9, 36.1] & \textbf{36.7} [36.2, 37.3] & 36.3 [35.8, 36.9] \\
    BLEU-1 &  39.7 [38.9, 40.5] & 40.5 [39.7, 41.3] & 39.3 [38.6, 40.2] \\
    BLEU-4  & 19.5 [19.0, 20.1] & 20.3 [19.8, 20.8] & 19.6 [19.1, 20.1] \\
    METEOR &  37.2 [36.6, 37.8] & \textbf{38.4} [37.8, 38.9] & 37.4 [36.9, 38.0] \\
    \midrule
    \multicolumn{3}{l}{\textit{\textbf{Clinical }}} \\
    RadGraph-F1  & 29.6 [29.0, 30.4] & \textbf{31.3} [30.7, 32.1] & \textbf{30.9} [30.1, 31.6] \\
    \rger   & 34.6 [34.0, 35.2] & \textbf{36.3 }[35.7, 37.0] & \textbf{35.7} [35.1, 36.5] \\
    CheXbert vector  & 44.1 [43.3, 45.2] & \textbf{45.9} [45.0, 46.8] & \textbf{46.4} [45.4, 47.3] \\
    RadCliQ (↓)  & 2.90 [2.86, 2.94] & \textbf{2.81} [2.77, 2.85] & \textbf{2.82} [2.78, 2.86] \\
    \macrofourteen   & 32.3 [30.8, 34.0] & 33.5 [32.1, 35.1] & \textbf{35.3} [33.5, 37.0] \\
    \microfourteen  & 46.0 [44.7, 47.3] & \textbf{49.2} [48.0, 50.3] & \textbf{50.5} [49.5, 52.0] \\
 \macromr*& 52.5 [50.0, 55.0] & 53.9 [51.5, 56.3] & 55.9 [53.6, 58.7] \\
 \micromr* & 55.6 [54.2, 57.0] & \textbf{59.9 }[58.6, 61.1] & \textbf{60.9 }[59.6, 62.5] \\
    RadFact/logical\_precision & 48.3 [47.1, 49.6] & \textbf{50.7} [49.5, 52.0] & \textbf{51.5} [50.2, 52.9] \\
    RadFact/logical\_recall & 42.8 [41.5, 43.9] & 44.4 [43.4, 45.6] & 43.4 [42.3, 44.6] \\
    RadFact/logical\_f1 & 45.4 [44.2, 46.5] & \textbf{47.4} [46.3, 48.6] & \textbf{47.1} [46.1, 48.3] \\
    \bottomrule
    \end{tabular}%
  \label{tab:maira2_test}%
\end{table*}%

\begin{figure*}[!h]
 % Caption and label go in the first argument and the figure contents
 % go in the second argument
\floatconts
  {fig:stratified_all}
  {\caption{Stratified F1-scores for the 14 CheXpert findings comparing the respective baselines, \ac{SoM} prompting and MAIRA-Seg models for (a) single view (b) multi-view experiments (c) Support for each pathological finding in the MIMIC-CXR test set.}} %\todo{Update stratified results for MAIRA-2-Seg}.}}
  {\includegraphics[width=\linewidth]{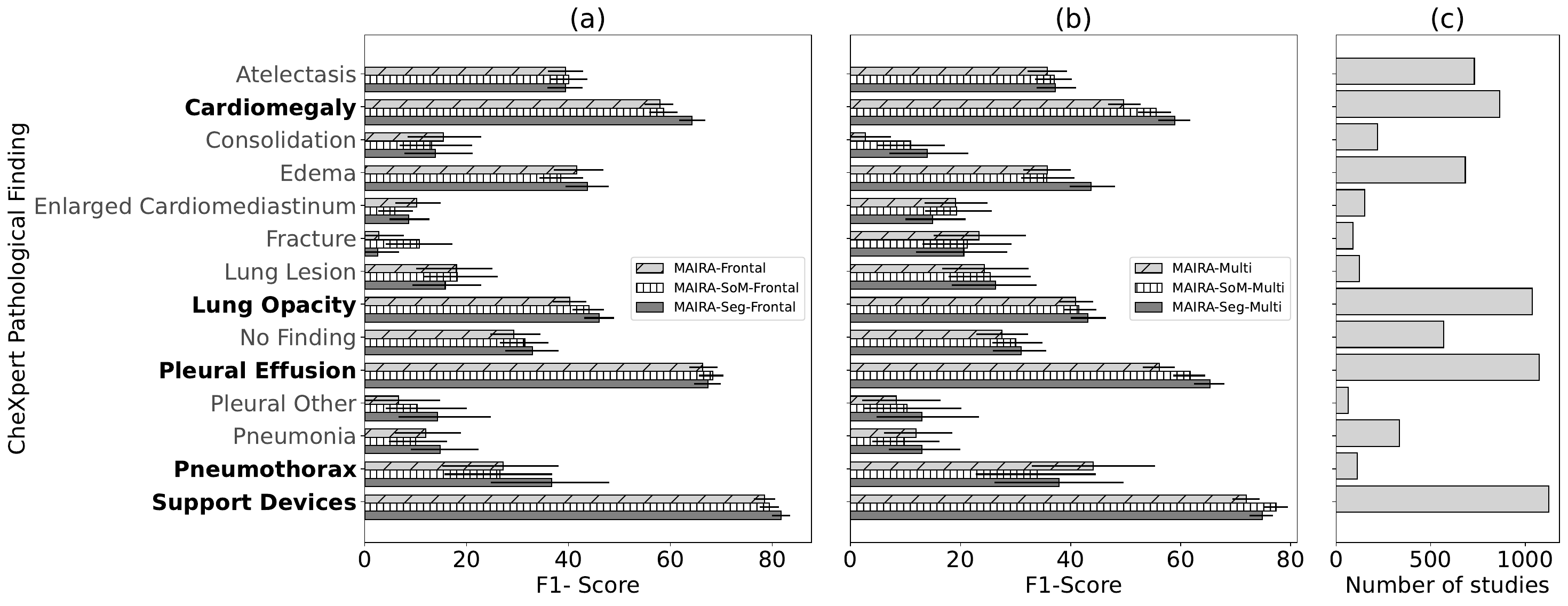}}
\end{figure*}

\section{Qualitative Results}\label{apd:second}
\label{app:qual}
We present qualitative results in \figureref{fig:qual1,fig:qual2} and \figureref{fig:qual4} for \mairasegone and \mairasegtwo, respectively. We present the target report and predicted findings of the baselines and corresponding proposed MAIRA-Seg methods. In the reports, we highlight the phrases selected for further analysis (gray) in the target findings, and the corresponding errors (red) and correct predictions (green). The image labelled as ``current image with mask pseudolabels'' is only for illustrating the mask pseudolabels used along with the CXR image in the MAIRA-Seg architecture, where the masks in green, red, yellow, blue, light-blue, white correspond to right lung, left lung, \ac{NGT}, \ac{CVC}, \ac{ETT}, and heart respectively.

%We find that..\todo{explain the qualitative figures}

In \figureref{fig:qual1}(a), we find that \mairaonebaseline hallucinates Cardiomegaly (``cardiac contours are mildly enlarged'') whereas this is correctly reported by \mairasegone. Also, bibasilar atelectasis is omitted by \mairaonebaseline but correctly stated by \mairasegone. Both \mairaonebaseline and \mairasegone report an \ac{NGT} (traced with yellow with \mairasegone prediction showing the corresponding \ac{NGT} mask pseudolabel). In \figureref{fig:qual1}(b), we note that \mairaonebaseline wrongly predicts the tip position of the IJ catheter (a type of \ac{CVC}), which is correctly predicted by \mairasegone (as correctly depicted by the blue \ac{CVC} mask). Moreover, \mairaonebaseline hallucinates atelectasis, and mentions `bilateral' pleural effusion -- this is correctly predicted as `right' only by \mairasegone (target report mentions `no left effusion'). In \figureref{fig:qual1}(c), we observe that again \mairaonebaseline erroneously mentions the tip of the IJ line in the right atrium, which is correctly predicted as cavo-atrial junction by \mairasegone, also depicted in the corresponding blue \ac{CVC} mask. We also find that \mairaonebaseline hallucinates cardiomegaly and omits right pleural effusion -- these are correctly predicted by \mairasegone. For \mairasegone, the extents of pleural effusions is mistaken as `moderate' compared to extensive, and atelectasis is omitted. In \figureref{fig:qual2}(a), \mairaonebaseline omits atelectasis and mentions of any support device. In contrast, \mairasegone correctly mentions atelectasis (although as `bibasilar' rather than `right'). \mairasegone mentions the two devices, namely, AICD and PICC line (a type of \ac{CVC}) correctly, and with the correct position for AICD, but erroneously for PICC line: interestingly, we note that the PICC line  mask is over-segmented because of overlap with AICD, and gives the appearance of `distal SVC' that is predicted by \mairasegone rather than mid-SVC. In \figureref{fig:qual2}(b), we observe that atelectasis is omitted by \mairaonebaseline but correctly mentioned by \mairasegone. In \figureref{fig:qual2}(c), \mairaonebaseline omits cardiomegaly and pacemaker position, correctly predicted by \mairasegone. The former wrongly mentions the side of the costophenic angle (this detail is not usually a part of the \textit{Findings} section). In \figureref{fig:qual4}(b), we note that \mairasegtwo correctly predicts the type, side and tip location of the dialysis catheter (corresponding to the blue mask) whereas \mairatwobaseline makes error in the approach. \mairasegtwo suggests moderate cardiomegaly and no pleural effusion, which are omitted and hallucinated respectively by \mairatwobaseline. \review{\mairatwobaseline doesn't mention the slightly low lung volumes at all, these are erroneously mentioned as `well expanded' by \mairasegtwo.}

\begin{figure*}[t]
\floatconts
  {fig:qual1}
  {\caption{Qualitative result for examples in the MIMIC-CXR test set, showing target and predicted \textit{Findings} using \mairaonebaseline and  \mairasegone. Mask pseudolabels are shown overlaid on the \ac{CXR} image for illustrative purposes (corresponding masks are used to obtain segmentation tokens).}}
  {
  \subfigure[][c]{\includegraphics[width=0.9\linewidth, clip, trim=40pt 9cm 15pt 35pt]{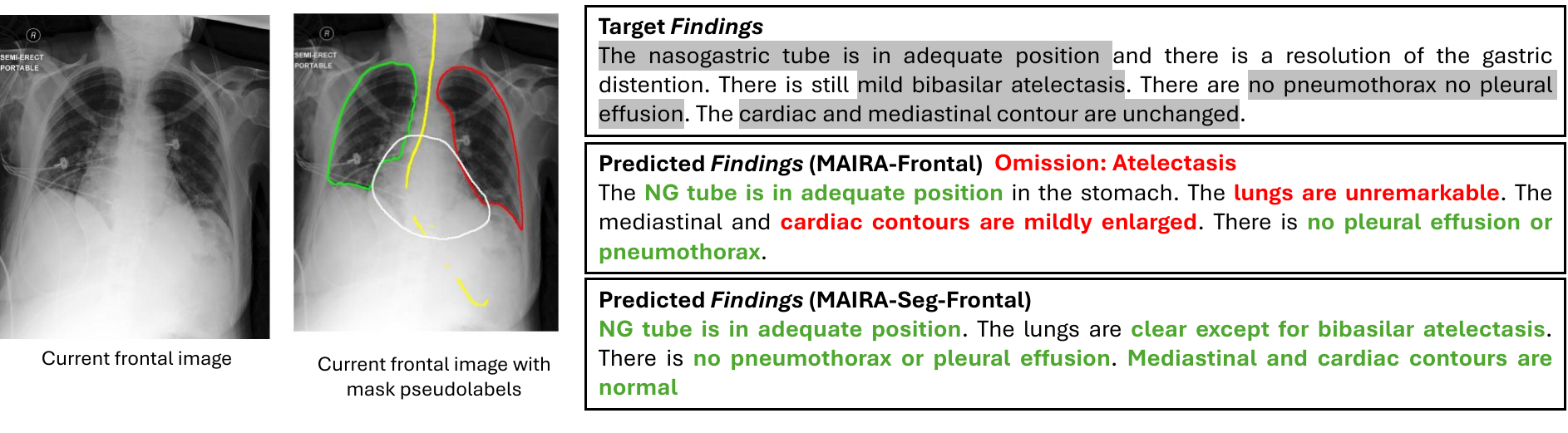}} 
    \qquad
    \subfigure[][c]{\includegraphics[width=0.9\linewidth, clip, trim=40pt 6cm 20pt 35pt]{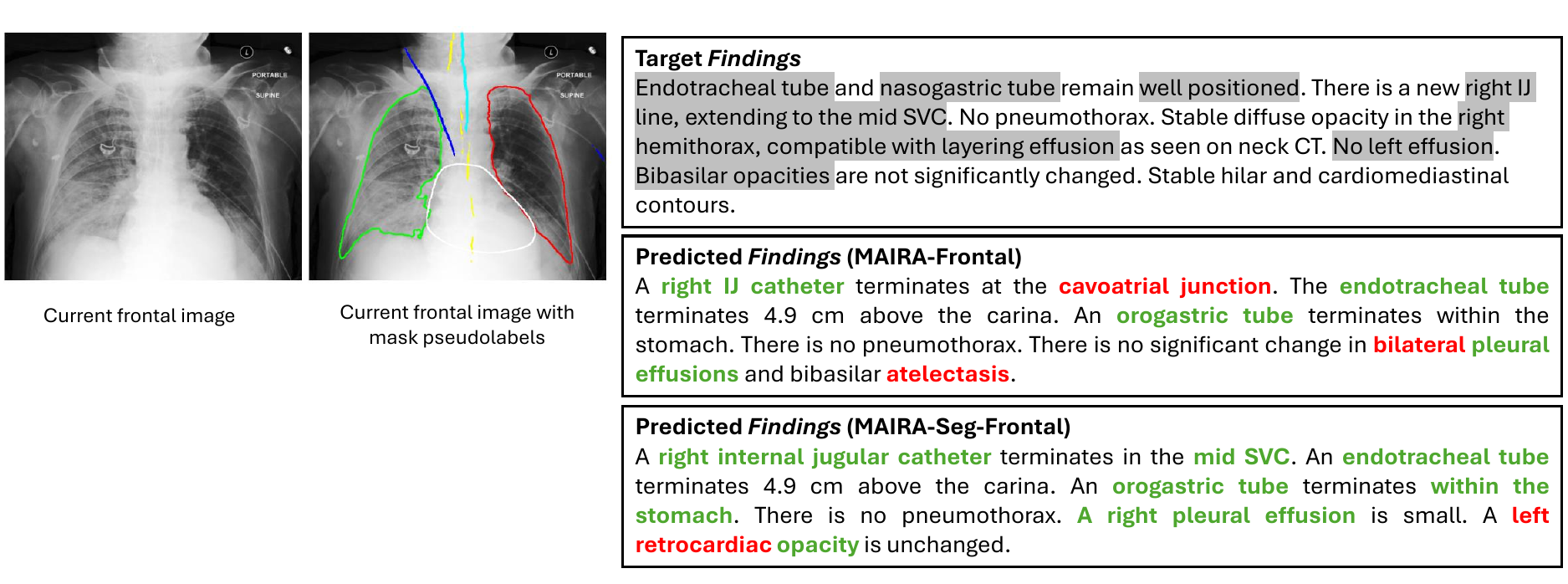}}
    \qquad
    \subfigure[][c]{\includegraphics[width=0.9\linewidth, clip, trim=40pt 6cm 20pt 35pt]{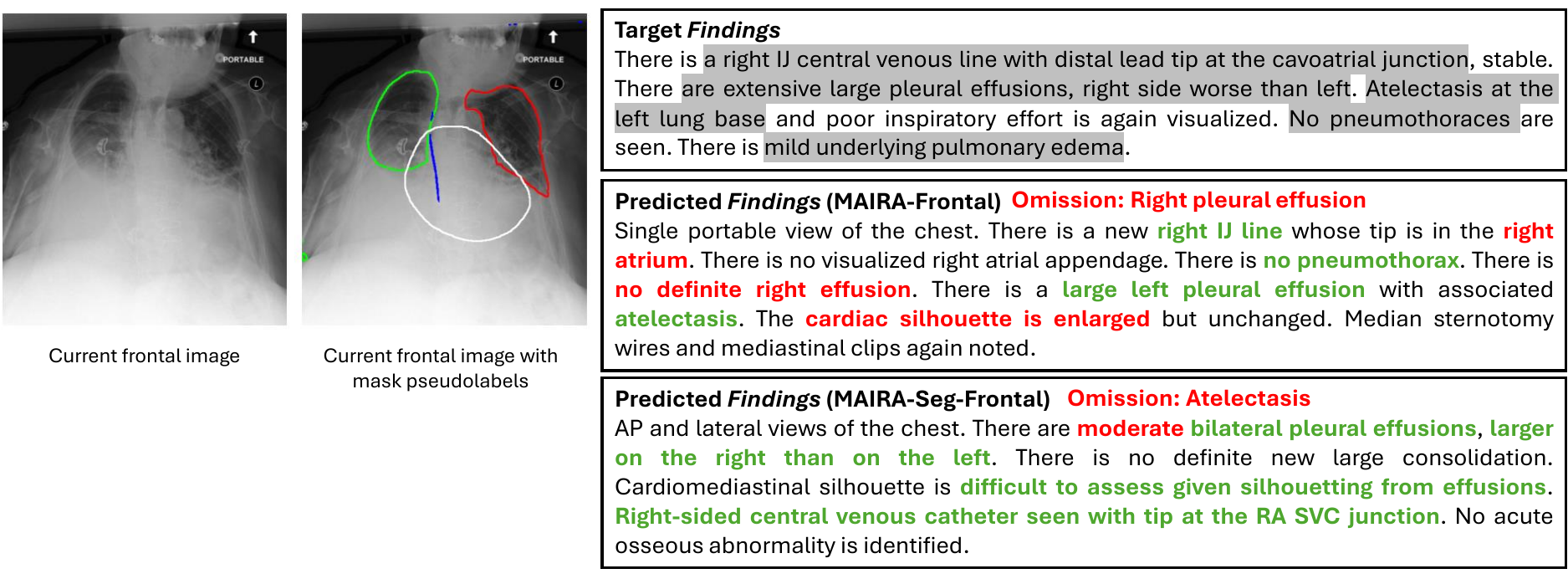}}
  }
\end{figure*}

\begin{figure*}[t]
\floatconts
  {fig:qual2}
  {\caption{(Contd.) Qualitative result for examples in the MIMIC-CXR test set, showing target and predicted \textit{Findings} using \mairaonebaseline and  \mairasegone. Mask pseudolabels are shown overlaid on the \ac{CXR} image for illustrative purposes (corresponding masks are used to obtain segmentation tokens).}}
  {
    \subfigure[][c]{\includegraphics[width=0.9\linewidth, clip, trim=50pt 6cm 5pt 35pt]{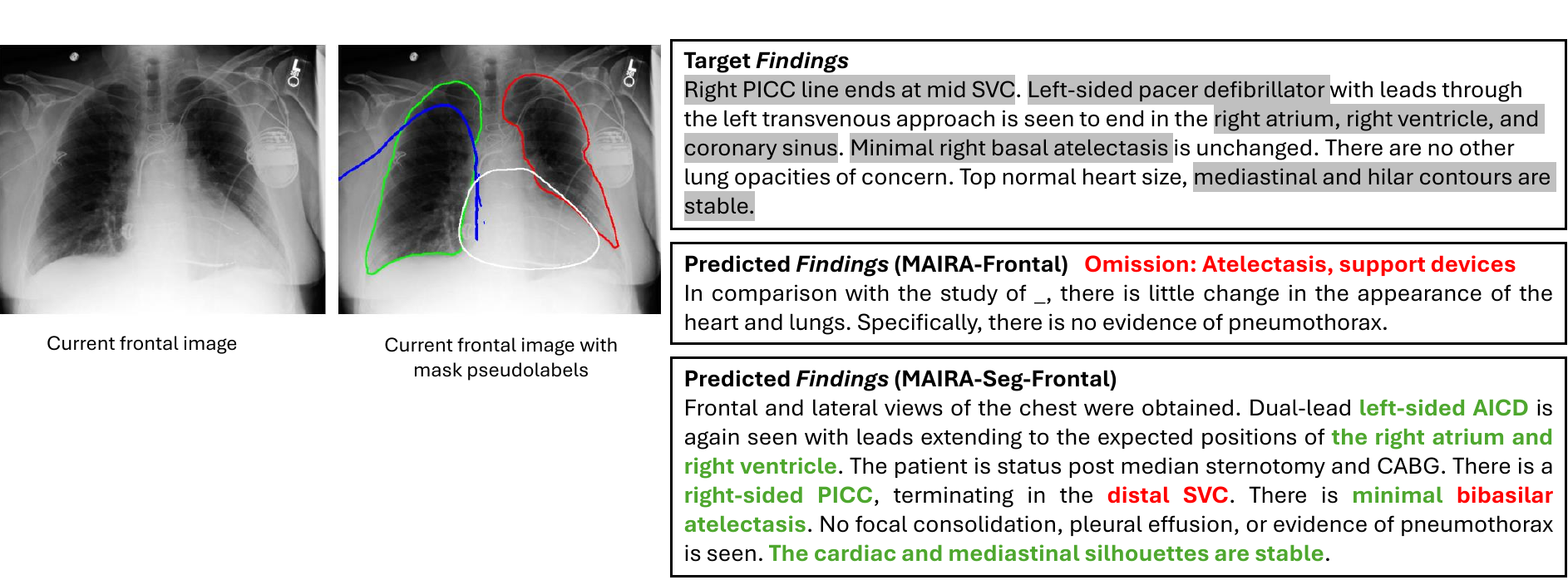}}
  \qquad
    \subfigure[][c]{\includegraphics[width=0.9\linewidth, clip, trim=40pt 7cm 15pt 35pt]{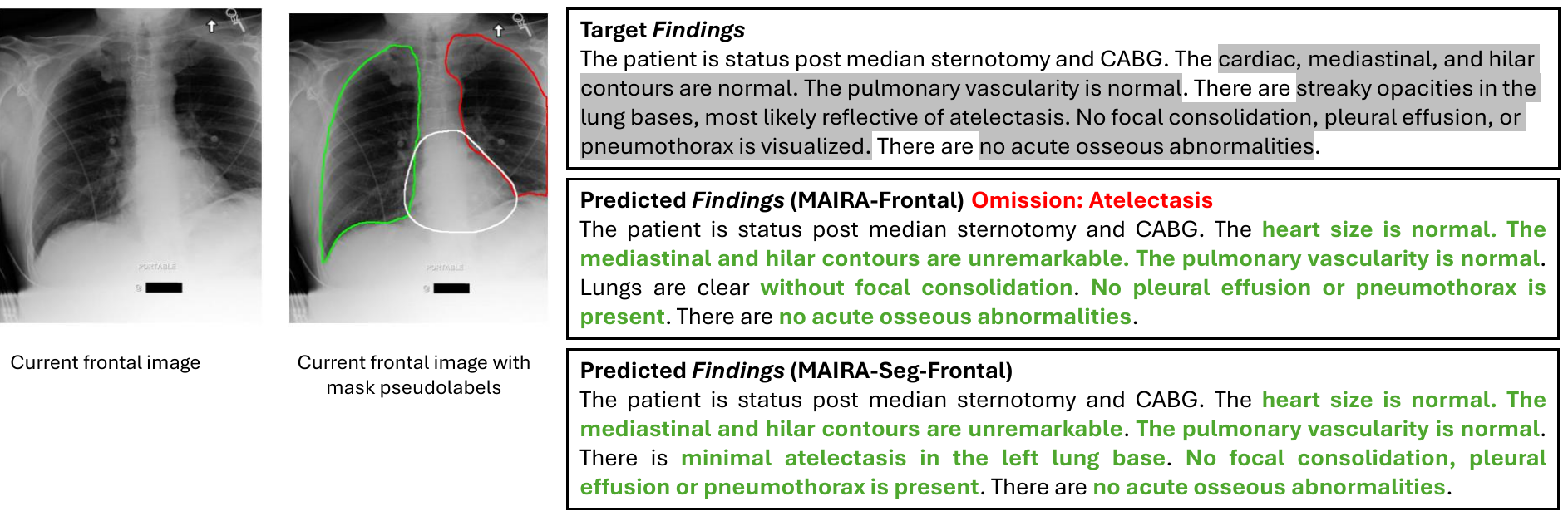}}
      \qquad
    \subfigure[][c]{\includegraphics[width=0.9\linewidth, clip, trim=40pt 4cm 20pt 35pt]{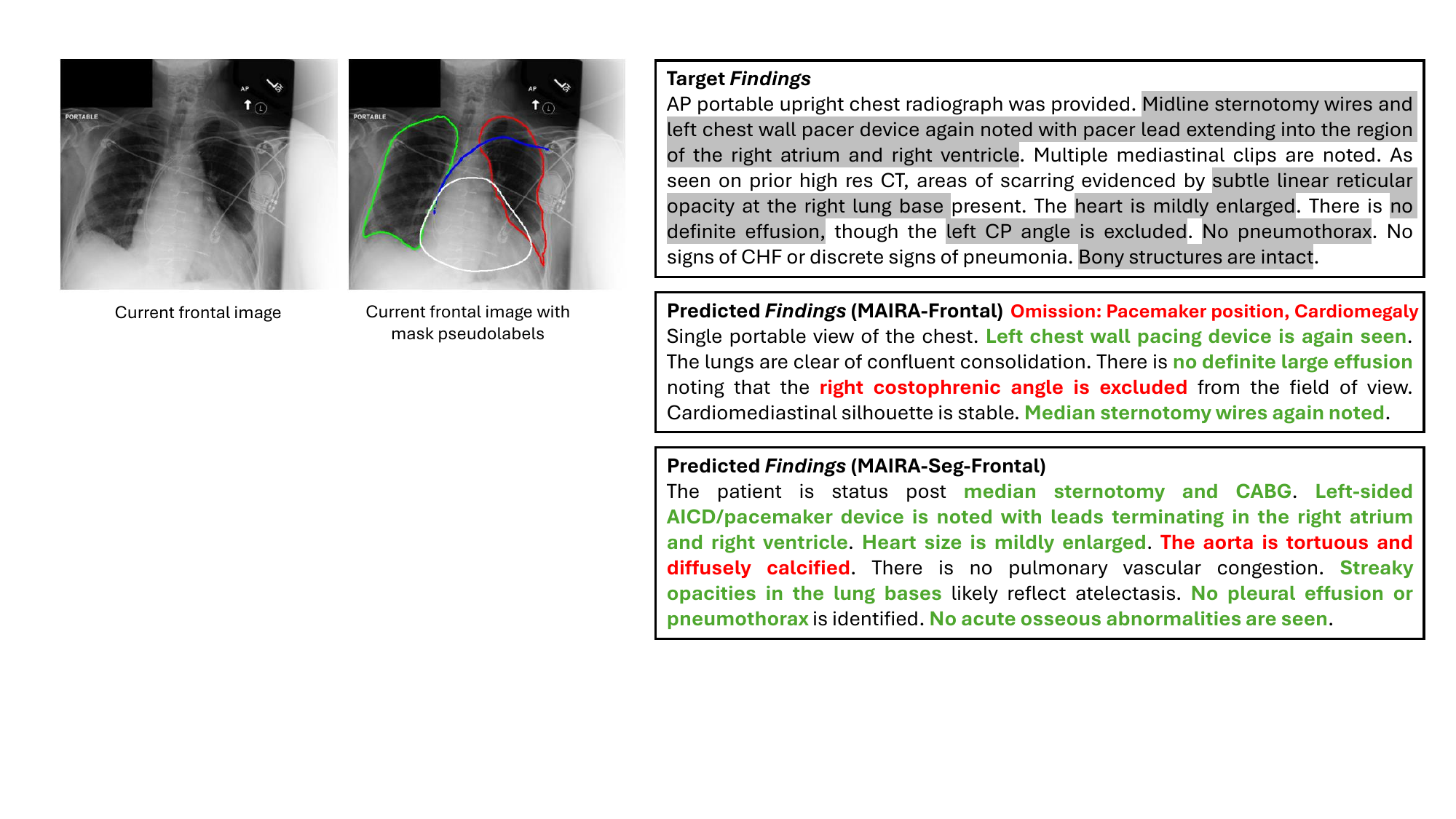}}
    }
\end{figure*}

\begin{figure*}[!h]
\floatconts
  {fig:qual4}
  {\caption{Qualitative result for examples in the MIMIC-CXR test set, showing target and predicted \textit{Findings} using \mairatwobaseline and \mairasegtwo. Mask pseudolabels are shown overlaid on the \ac{CXR} image for illustrative purposes (corresponding masks are used to obtain segmentation tokens).}}
  {
    \subfigure[][c]{\includegraphics[width=0.9\linewidth, clip, trim=30pt 3cm 90pt 40pt]{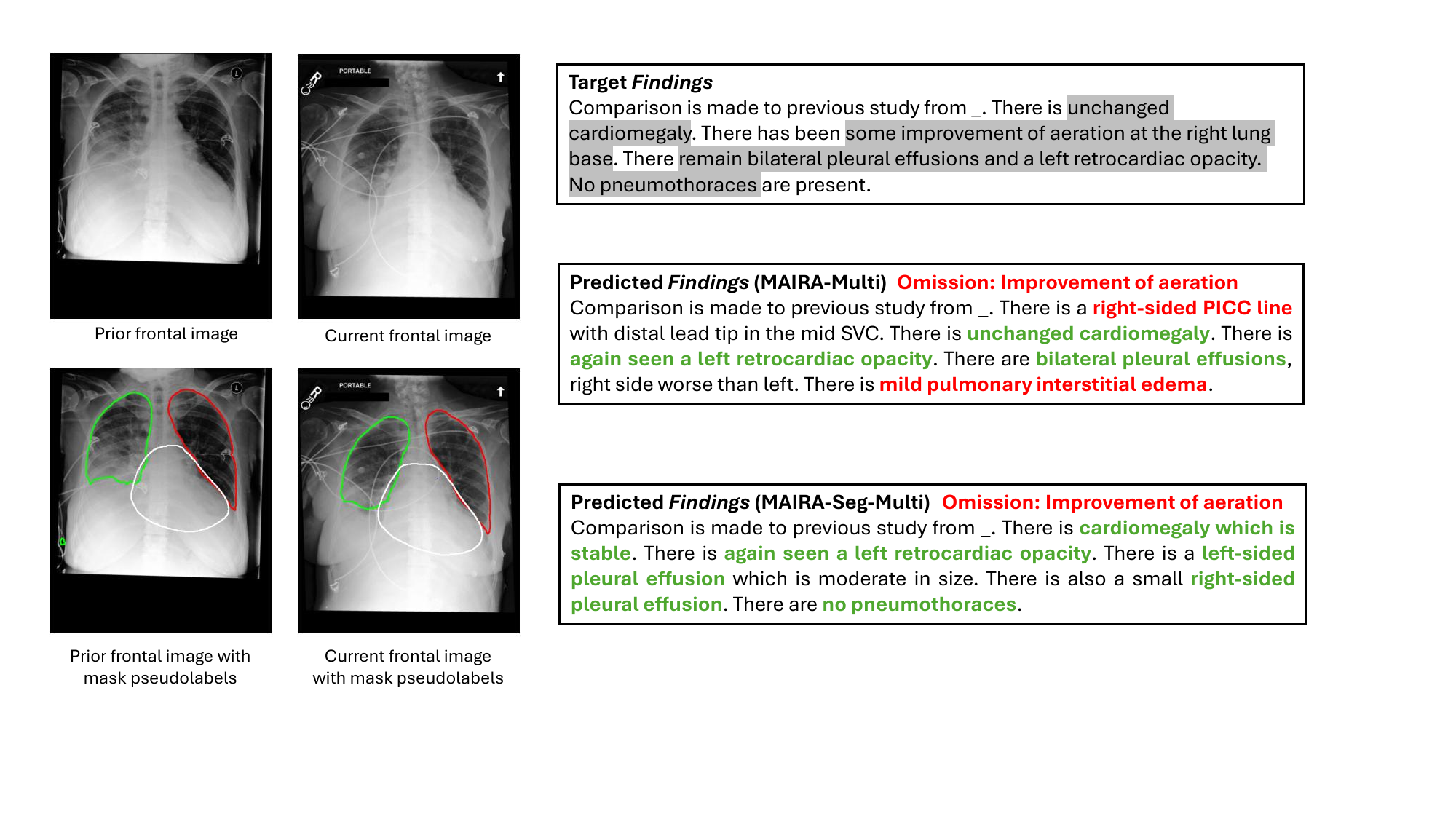}}
  \qquad
    \subfigure[][c]{\includegraphics[width=0.9\linewidth, clip, trim=30pt 5cm 20pt 10pt]{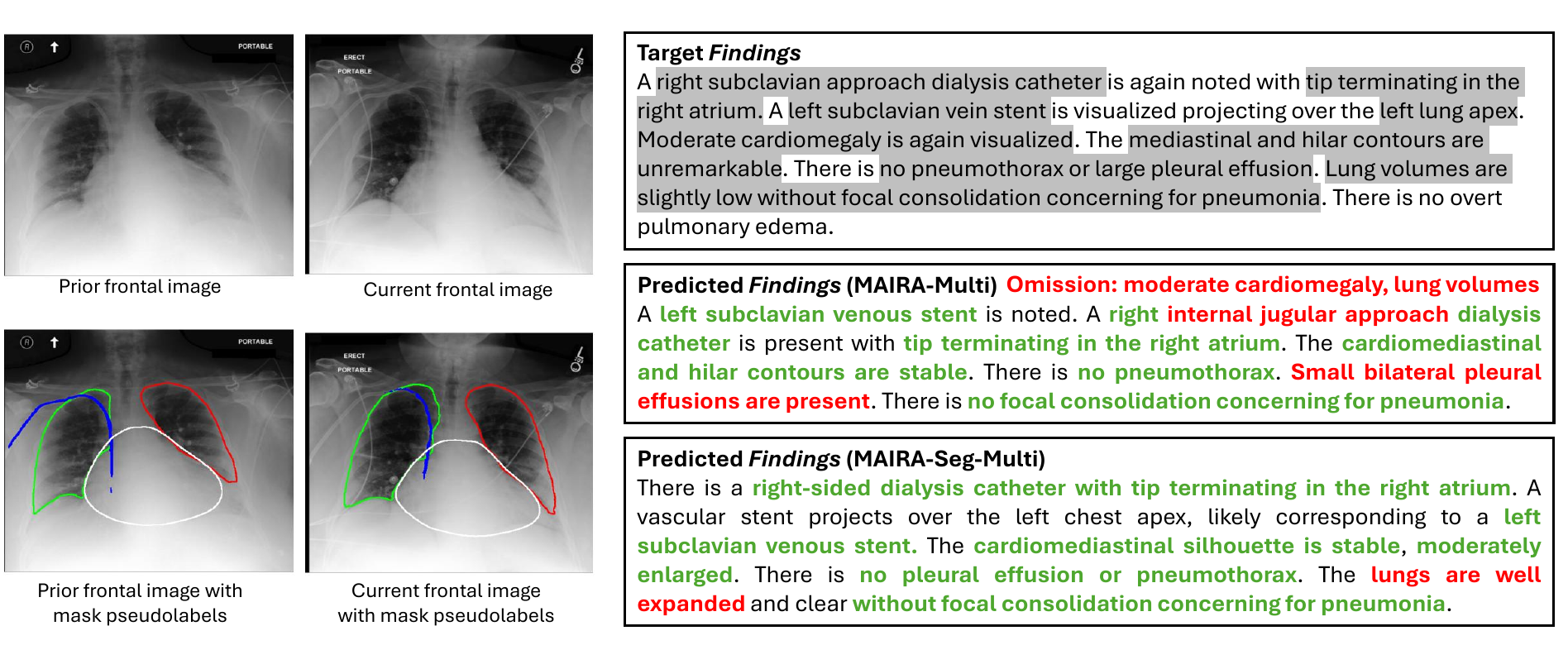}}
    }
\end{figure*}

\section{Implementation Details}\label{apd:third}

We train MAIRA-Seg models (\mairasegone and \mairasegtwo) with a conventional autoregressive cross-entropy loss on the MIMIC-CXR training set. Following the training method in~\citet{hyland2024maira1}, we do a single stage of training with a frozen image encoder and trainable adapter and \ac{LLM}. We train for 3 epochs and use the final checkpoint in evaluations. We use the AdamW optimiser and a cosine learning rate scheduler. We train \mairasegone on 4 NVIDIA A100 GPUs with a global training batch size of 128, training time $\approx$15h, warm-up of 0.03, and a learning rate of \num{2e-5}. We train \mairasegtwo across 16 NVIDIA A100 GPUs with a global training batch size of 256,  training time $\approx$11h, warmup ratio 0.03, and a learning rate of \num{2e-5}. We use a pre-trained \raddino image encoder~\citep{pérezgarcía2024raddinoexploringscalablemedical} using the same weights as used in~\citet{bannur2024maira2groundedradiologyreport}.

\review{We train expert segmentation models on 8 NVIDIA V100 GPUs. We use a training batch size 80 (10 per GPU), Adam optimizer, base learning rate \num{5e-4}, and a cosine learning rate scheduler. We use the following preprocessing and augmentations: centre-cropping and resizing (512 $\times$ 512), random horizontal flip (except left–right lungs), random affine transform, elastic transform, random brightness and contrast jittering, and random gamma adjustments. The segmentation models are trained for 100 epochs. The checkpoint with minimum loss on the validation set is used for inference on the test set. We use a 70/15/15 split by subjects for train, validation and test sets, respectively, and report metrics on the test set.}

\end{document}

%% file: macros.tex
\newcommand{\domainimageencoder}{R\textsc{ad}-DINO\xspace}
\newcommand{\maira}{MAIRA-1\xspace}
\newcommand{\mairatwo}{MAIRA-2\xspace}
\newcommand{\raddino}{\domainimageencoder}
\newcommand{\modelname}[1]{\texttt{#1}}

\newcommand{\reportsection}[1]{\textit{#1}}
\newcommand{\findings}{\reportsection{Findings}}
\newcommand{\impression}{\reportsection{Impression}}
\newcommand{\indication}{\reportsection{Indication}}
\newcommand{\todo}[1]{\textcolor{purple}{[TODO: #1]}}
\newcommand{\review}[1]{\textcolor{black}{#1}}

% Adapted from https://tex.stackexchange.com/a/312583
\newcommand{\ctext}[2]{{\sethlcolor{#1}\hl{#2}}}
\newcommand{\reporterror}[1]{\ctext{red!20}{#1}}
\newcommand{\reportomission}[1]{\ctext{lightgray!50}{[\textit{#1}]}}
\newcommand{\reporthighlight}[1]{\ctext{ProcessBlue!20}{#1}}

\newcommand{\mairaseg}{MAIRA-Seg\xspace}
\newcommand{\mairasegone}{MAIRA-Seg-Frontal\xspace}
\newcommand{\mairasegtwo}{MAIRA-Seg-Multi\xspace}
\newcommand{\mairaonebaseline}{MAIRA-Frontal\xspace}
\newcommand{\mairatwobaseline}{MAIRA-Multi\xspace}
\newcommand{\somone}{MAIRA-SoM-Frontal\xspace}
\newcommand{\somtwo}{MAIRA-SoM-Multi\xspace}
\newcommand{\extractor}{segmentation tokens extractor\xspace}

\newcommand{\rger}{RG\textsubscript{ER}\xspace}
\newcommand{\macrofourteen}{Macro F$_1$-14\xspace}
\newcommand{\microfourteen}{Micro F$_1$-14\xspace}
\newcommand{\macromr}{Macro F$_1$-MR\xspace}
\newcommand{\micromr}{Micro F$_1$-MR\xspace}

%% file: acronyms.tex
\begin{acronym}
    \acro{AI}{Artificial Intelligence}
    \acro{AUROC}{area under the receiver operating characteristic curve}
    \acro{BMI}{body mass index}
    \acro{CI}{confidence interval}
    \acro{CXR}{chest X-ray}
    \acro{LLM}{large language model}
    \acro{MIM}{masked image modelling}
    \acro{PHI}{protected health information}
    \acro{PTX}{pneumothorax}
    \acro{SSL}{self-supervised learning}
    \acro{SOTA}{state-of-the-art}
    \acro{ViT}{vision transformer} 
    \acro{VQA}{Visual Question Answering}
    \acro{CVC}{Central Venous Catheter}
    \acro{SGC}{Swan-Ganz Catheter}
    \acro{ETT}{Endo-Tracheal Tube}
    \acro{NGT}{Naso-Gastric Tube}
    \acro{MLLM}{multi-modal large language model}
    \acro{MLP}{multi-layer perceptron}
    \acro{SoM}{set-of-marks}
    \acro{MAIRA}{Multimodal AI for Radiology Applications}
    \acro{CNN}{convolutional neural network}
    \acro{ROI}{region of interest}
    \acroplural{ROI}{regions of interest}
    \acro{SAM}{Segment Anything Model}
\end{acronym}